\documentclass[10pt,journal,compsoc]{IEEEtran}

\ifCLASSOPTIONcompsoc
  \usepackage[nocompress]{cite}
\else
  \usepackage{cite}
\fi

\usepackage{tabularx}
\usepackage{multirow} 
\usepackage{amssymb}
\usepackage{color}
\usepackage{graphicx}
\usepackage{textcomp}
\usepackage{stfloats}
\usepackage{url}
\usepackage{verbatim}
\usepackage{amsmath,amsfonts}
\usepackage{algorithmic}
\usepackage{algorithm}
\usepackage{array}
\ifCLASSINFOpdf
\else
\fi
\begin{document}
%
\title{PSDF: Prior-Driven Neural Implicit Surface Learning 
	for Multi-view Reconstruction}
%
%
%
%

\author{Wanjuan~Su,~
        Chen~Zhang,~
        Qingshan~Xu,~
        and~Wenbing~Tao
\IEEEcompsocitemizethanks{\IEEEcompsocthanksitem W. Su, C. Zhang and W. Tao are with the School of Artificial Intelligence and Automation, Huazhong University of Science and Technology, Wuhan 430074, China. E-mail: \{suwanjuan, zhangchen\_, wenbingtao\}@hust.edu.cn\protect\\
Q. Xu is with the School of Computer Science and Engineer, Nanyang Technological University, 50 Nanyang Ave, 639798, Singapore. E-mail: qingshan.xu@ntu.edu.sg 
\IEEEcompsocthanksitem Corresponding Author: Wenbing Tao.}
\thanks{Manuscript received April 19, 2005; revised August 26, 2015.}}

%
%

\markboth{Journal of \LaTeX\ Class Files,~Vol.~14, No.~8, August~2015}%
{Shell \MakeLowercase{\textit{et al.}}: Bare Advanced Demo of IEEEtran.cls for IEEE Computer Society Journals}
%



\IEEEtitleabstractindextext{%
\begin{abstract}
Surface reconstruction has traditionally relied on the Multi-View Stereo (MVS)-based pipeline, which often suffers from noisy and incomplete geometry. This is due to that although MVS has been proven to be an effective way to recover the geometry of the scenes, especially for locally detailed areas with rich textures, it struggles to deal with areas with low texture and large variations of illumination where the photometric consistency is unreliable.
Recently, Neural Implicit Surface Reconstruction (NISR) combines surface rendering and volume rendering techniques and bypasses the MVS as an intermediate step, which has emerged as a promising alternative to overcome the limitations of traditional pipelines.
While NISR has shown impressive results on simple scenes, it remains challenging to recover delicate geometry from uncontrolled real-world scenes which is caused by its underconstrained optimization. To this end, the framework PSDF is proposed which resorts to external geometric priors from a pretrained MVS network and internal geometric priors inherent in the NISR model to facilitate high-quality neural implicit surface learning. Specifically, the visibility-aware feature consistency loss and depth prior-assisted sampling based on external geometric priors are introduced. These proposals provide powerfully geometric consistency constraints and aid in locating surface intersection points, thereby significantly improving the accuracy and delicate reconstruction of NISR. Meanwhile, the internal prior-guided importance rendering is presented to enhance the fidelity of the reconstructed surface mesh by mitigating the biased rendering issue in NISR. Extensive experiments on the Tanks and Temples dataset show that PSDF achieves state-of-the-art performance on complex uncontrolled scenes.
\end{abstract}

}

\maketitle

\IEEEdisplaynontitleabstractindextext

%
\IEEEpeerreviewmaketitle

\ifCLASSOPTIONcompsoc
\IEEEraisesectionheading{\section{Introduction}\label{sec:introduction}}
\else
\section{Introduction}
\fi
\IEEEPARstart{S}{urface} reconstruction from posed multi-view images is one of the fundamental problems in 3D computer vision \cite{Li2022, Petrov2023}.
Traditional Multi-View Stereo (MVS)-based pipelines typically utilize depth maps or point clouds derived from MVS \cite{Schoenberger2016, Wei2022} as intermediate representations to reconstruct the scene's triangle mesh.
As MVS generally treats 3D reconstruction as a correspondence-matching problem, it demonstrates an advantage in recovering geometry from areas with rich textures.
However, as the core clue, i.e., photometric consistency, for correspondence-matching is not hold in low-texture and large illumination variation areas, the geometry recovered by MVS always suffers from noise or incomplete problems, leading to the same problems on the reconstructed surface meshes \cite{Zhang2022}.
Neural Implicit Surface Reconstruction (NISR) methods \cite{Wang2021, Yariv2021} have recently garnered great attention, which directly reconstruct surface geometry from multi-view images based on differentiable surface rendering and volume rendering techniques \cite{Yariv2020, Mildenhall2020}, thereby avoiding the accumulated errors caused by MVS as an intermediate step.

Although existing NISR methods have shown promising results in simple scenes, they encounter difficulties in recovering high-quality geometry from uncontrolled real-world scenes \cite{Zhang2022}. 
This challenge arises from the heavy reliance on color loss for optimization, which primarily focuses on color field reconstruction, treating the geometry field reconstruction as a byproduct. As a result, vanilla NISR methods tend to only recover the global structure lacking delicate details.
To enhance the optimization of geometric fields, priors are introduced by some methods which can be mainly divided into two kinds. One using monocular priors to constrain the optimization of textureless regions \cite{Yu2022, Wang2022a}, and another either introducing explicit geometric constraints to constrain the learning of Signed Distance Functions (SDFs) as the MVS doses or directly build NISR upon the MVS  \cite{Fu2022, Zhang2021, Darmon2022, Zhang2022, Chen2023, Zhang2023}.
However, the former shows poor geometric accuracy due to the lack of geometric consistency constraints, while for the latter, the photoconsistency constraint suffers from limited representative ability of hand-crafted similarity metrics \cite{Fu2022, Darmon2022} and erroneous constraints for occluded pixels \cite{Zhang2021, Chen2023}, the strict supervision of SDFs based on the depth priors obtained from MVS inevitably suffers from the noisy in the priors \cite{Zhang2021, Fu2022, Zhang2022, Zhang2023}.
As a result, their performance remains challenged in dealing with uncontrolled real-world scenes.

To this end, we thoroughly explore the utilization of the external priors for powerful geometric constrain from a pretained MVS network \cite{Yao2018} which can provide valuable clues such as visibility information, discriminative features, and depths.
Firstly, the visibility-aware feature consistency loss is introduced to provide a robust photoconsistency constraint for optimization when recovering unconstrained scenes. It leverages the deep features and visibility maps extracted from MVS, which can provide stronger similarity measure from discriminative features and proper occlusion handling using the visibility information thus avoiding the limitations in existing methods \cite{Fu2022, Chen2023}.
Then, the depth information estimated by the MVS is further exploited to promote neural implicit surface learning. Instead of imposing strictly geometric constraints based on the depth prior like \cite{Zhang2021, Fu2022, Zhang2022, Zhang2023} which is inevitably affected by the hard-to-filter noise, we propose the depth prior-assisted sampling for supplemental samples generating. In this way, we can prevent the adverse effects caused by the aforementioned factor, and also utilize depth priors for more delicate reconstruction and locating the surface intersection points efficiently.

Furthermore, the biased surface rendering caused by the volume rendering integral is another problem that needs to be solved.
Although explicit geometric consistency constraints have shown to be effective to mitigate this \cite{Fu2022}, it persists due to the inherent nature of volume rendering and hinders the high-fidelity surface rendering in complex scenes.
To enhance the fidelity of surface reconstruction, the inherent knowledge within the NISR is further explored.  
Note that the densely distributed near-surface points sampled by the hierarchical sampling algorithm  \cite{Wang2021} or error-bounded sampling algorithm \cite{Yariv2021} can be used for linear interpolation to calculate the point at the  zero-level set of learned SDFs.
Additionally, driven by volume rendering, the rendered depth can also indicate the position of the surface intersection point.
Based on these observations, we introduce the internal prior-guided importance rendering to direct the network's attention toward the surface points that uses the two aforementioned key points for volume rendering to further mitigate biased rendering effects.
In this case, the total training process can be divided into non-importance rendering and importance rendering. The former use the densely distributed near-surface points for rendering with the utilization of external priors, the latter leverages the internal prior that derives the two informed points from the former stage towards unbiased rendering.

By integrating the aforementioned two rendering stages into NISR, the prior-driven neural implicit surface learning for multi-view reconstruction is formed, named PSDF, which thoroughly explores and exploits the external and internal priors for high-quality surface reconstruction,  even in unconstrained scenes. Extensive experiments on Tanks and Temples \cite{Knapitsch2017} and DTU \cite{Aanaes2016} datasets validate that PSDF excels in both controlled and uncontrolled scenes, outperforming existing methods and achieving state-of-the-art results on the Tanks and Temples dataset. Notably, on the training set of Tanks and Temples \cite{Knapitsch2017} dataset, PSDF shows impressive improvements of 357.63$\%$, 77.04$\%$, 53.37$\%$ and 41.26$\%$ compared with VolSDF \cite{Yariv2021}, MonoSDF \cite{Yu2022}, Geo-Neus \cite{Fu2022} and Neus \cite{Wang2021}, respectively.
To summarize, our contributions are as follows:
\begin{itemize}
	\item We propose PSDF, a prior-driven neural implicit surface learning method for multi-view reconstruction, which exploits both external and internal geometric priors for high-quality surface reconstruction; 
	\item The visibility-aware feature consistency loss and depth prior-assisted sampling are introduced, both derived from external geometric priors obtained from a pre-trained MVS network, which can improve the precision of recovered surfaces significantly;
	\item To mitigate the biased surface rendering induced by volume rendering and ensure high-fidelity surface reconstruction, the internal prior-guided importance rendering is presented by considering internal geometric cues.
\end{itemize}

\section{Related Works}
\label{sec:related_works}

\subsection{Multi-view Surface Reconstruction} 
Traditional multi-view surface reconstruction methods often transform dense representations obtained from MVS into surface meshes using techniques like  Delaunay triangulation \cite{Vu2011}, screened Poisson Surface Reconstruction (sPSR) \cite{Kazhdan2013} and TSDF fusion \cite{Curless1996}. The quality of the resulting meshes is closely linked to the results of the MVS.
The core principle of MVS dictates that a surface point must exhibit photometric consistency across all visible views. 
While traditional MVS methods \cite{Schoenberger2016} rely on manually engineered similarity metrics to establish dense correspondences, learning-based MVS methods \cite{Yao2018, Gu2020, Su2023} utilize powerful deep features to achieve robust correspondence matching.
However, the challenge arises in regions with low textures and varying illumination, where photometric consistency becomes ambiguous. Consequently, even advanced learning-based MVS methods \cite{Zhang2023a} result in noisy and incomplete reconstructions. 

Recent advances in neural implicit methods \cite{Niemeyer2020, Yariv2020} have diverged from using MVS as intermediate representations, and instead directly learn implicit surface representations from multi-view images using Multi-Layer Perceptrons (MLPs). This approach has the potential to address the limitations of classic MVS-based pipelines.
NeRF \cite{Mildenhall2020} uses a density-based volume rendering technique with MLPs to represent 3D scenes implicitly, where novel views can be rendered.
Subsequently, several methods \cite{Oechsle2021, Wang2021, Yariv2021} incorporate volume rendering with surface rendering to learn implicit surface representations without the need of masks for supervision.

Very recently, several variants based on Neus \cite{Wang2021} and VolSDF \cite{Yariv2021} are proposed.
NeuralWarp \cite{Darmon2022} proposes patch warping to refine the neural implicit
surfaces geometry in a post-processing manner.
Instant-NGP \cite{Mueller2022} introduced grid-based multi-resolution hash encoding, showing an impressive ability to learn high-frequency details.  Thus, the hash encoding is introduced in some methods, e.g., MonoSDF \cite{Yu2022}, NeuDA \cite{Cai2023} and PermutoSDF \cite{Rosu2023}.
Despite these advancements, these methods either focus on object reconstruction from simple scenes or exhibit limitations in handling large-scale complex scenes.
Neuralangelo \cite{Li2023} introduces the hash encoding to enhance the representation ability of the network, and the numerical gradients and coarse-to-fine optimization are proposed to improve the quality of surface reconstruction.

\begin{figure*}[htbp]
	\setlength{\abovecaptionskip}{1.5mm}
	\setlength{\belowcaptionskip}{0cm}
	\center
	\includegraphics[width=0.98\textwidth]{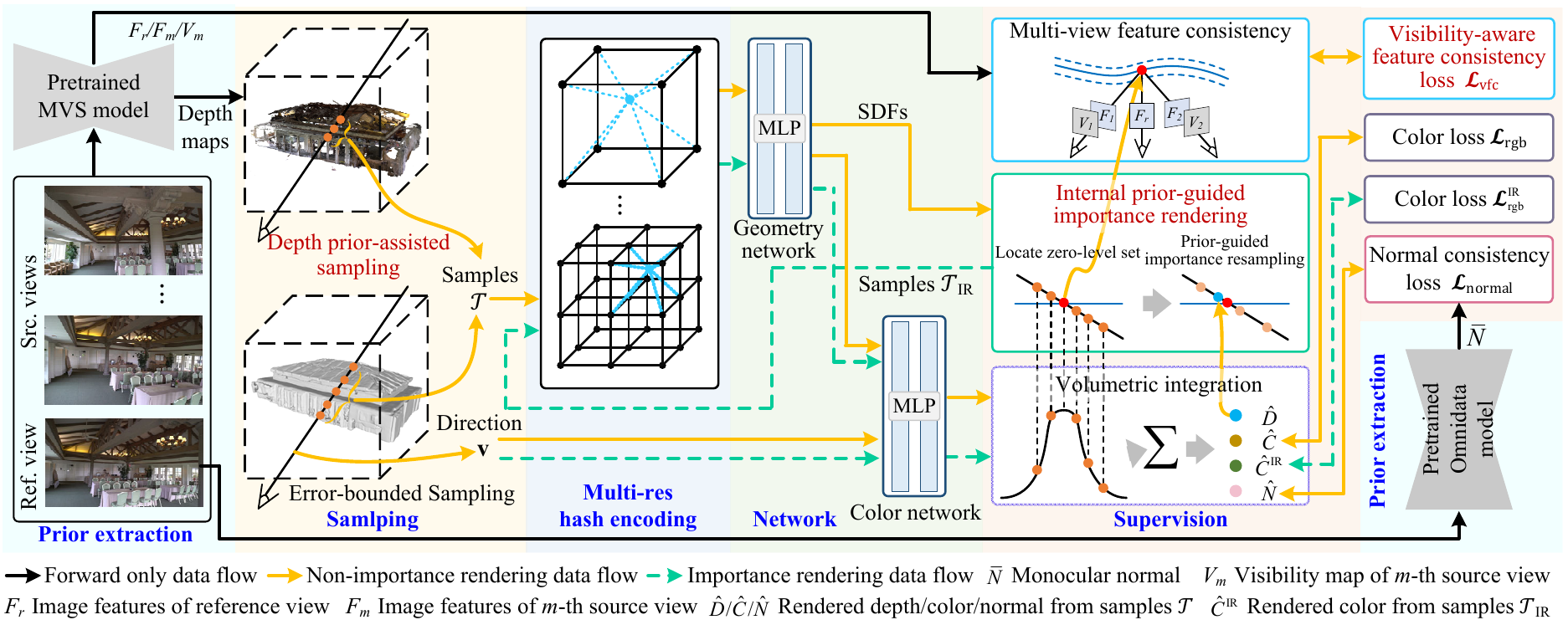}
	\caption{The overview of PSDF. The training of PSDF is divided into two stages: non-importance rendering and importance rendering. For non-importance rendering stage, the samples $\mathcal{T}$ are obtained by error-bounded sampling and depth prior-assisted sampling, and the training is supervised by the external geometric priors. For importance rendering, the internal priors  derived from non-importance rendering stage are used to form importance samples $\mathcal{T}_{\textrm{IR}}$ for rendering.}
	\label{framework}
\end{figure*}

\subsection{Priors-guided Neural Implicit Surface Reconstruction}
Vanilla NISR methods \cite{Oechsle2021, Wang2021, Yariv2021} primarily rely on the color loss to model the color field, which indirectly informs the geometry field. However, this approach faces challenges in low-textured regions where the implicit photometric consistency assumption of volume rendering is not valid, and there is biased surface rendering caused by the volume rendering integral.
Some methods aim to handle the photoconsistency ambiguity problem of low-texture areas in indoor scenes. Manhattan-SDF \cite{Guo2022} incorporates the Manhattan-world assumption with semantic constraints in the optimization process, and depth priors obtained from MVS are further utilized as supervision. 
Monocular normal and depth priors are introduced in MonoSDF \cite{Yu2022} for supervision to improve the performance of textureless and less-observed areas.

\vspace{-0.12cm}
Other methods focus on addressing the biased surface rendering caused by volume rendering. 
For instance, Geo-Neus \cite{Fu2022} employs explicit geometry optimization on the location of zero-level set of SDFs by utilizing sparse points from structure from motion and Normalization Cross Correlation (NCC)-based photometric consistency. Similarly, D-Neus \cite{Chen2023} incorporates geometry bias loss and multi-view feature consistency loss to achieve similar objectives.
While these explicit supervisions can enhance the geometric consistency of the reconstructed surfaces, they suffer from either the non-discriminative representation ability of hand-crafted similarity measure or the occlusion problem. 
Zhang et al. \cite{Zhang2023} proposes a new SDF-induced density function for unbiasd rendering and geometry priors from a pre-trained MVS network are adopted for the zero-level set supervision.
MVSDF \cite{Zhang2021} and RegSDF \cite{Zhang2022} build NISR upon the pre-trained MVS network, the former proposes the depth-based distance loss and multi-view feature consistency loss, the latter proposes the Hessian regularization and minimal surface regularization for optimization.
These methods impose strong geometric supervision on NISR based on the priors of MVS, making the recovered mesh susceptible to the adverse effects of noise in priors. Furthermore, the performance of NISR is limited by the quality of the point cloud.

\section{Method}
Given multi-view images of a scene with their camera parameters, we aim to recover its geometry surface. 
The pipeline of the proposed PSDF is illustrated in Fig. \ref{framework}. As shown, the training process of PSDF is divided into non-importance rendering and importance rendering. 
In the non-importance rendering stage, the external geometric priors derived from the pretrained MVS network and Omnidata model \cite{Eftekhar2021} are used for efficiently locating the surface intersection points and  guiding the learning process of the geometry field. This integration of prior knowledge helps to enhance the accuracy of the geometry reconstruction. 
In the importance rendering stage, we utilize the prior knowledge acquired during the non-importance rendering stage to mitigate biased surface rendering.

This section will describe the proposed PSDF in detail.
We first introduce the preliminaries for PSDF in Section \ref{sec:preliminaries}, which presents the general framework of NISR \cite{Yariv2021} and multi-res hash encoding \cite{Mueller2022}; then, we will introduce the proposed external geometric prior-guided learning
in Section \ref{sec:external_geometric_prior-guided_learning} which includes visibility-aware feature consistency loss and depth prior-assisted sampling; next, the proposed internal prior-guided importance rendering is presented in Section \ref{sec:internal_prior-guided_importance_rendering}; finally, the details of the supervision losses used for PSDF are given in Section \ref{sec:optimization}.

\subsection{Preliminaries}
\label{sec:preliminaries}

\subsubsection{Neural Implicit Surface Reconstruction}
\label{sec:neural_implicit_surface_reconstruction}
NISR \cite{Yariv2021} parameterizes the scene as a geometry field and a color field. The geometry field is described by SDFs which are modeled by a geometry network consisting of MLPs $f_\theta$ with learnable parameters $\theta$. 
The zero-level set of these SDFs defines the surface $\mathcal{S}$ as $\mathcal{S} = \{\mathbf{p} \in \mathbb{R}^3|f_\theta(\mathbf{p}) = 0\}$, where $\mathbf{p}$ represents the 3D position being queried.
To link the geometry field with the color field for volume rendering, the SDF-induced density function is introduced. Specifically, given a pixel $x$ for example, a ray $\mathbf{r}$ is cast from the camera center $\mathbf{o}$ through the pixel $x$ along its view direction $\mathbf{v}$, $P$ points $\mathbf{p}_\mathbf{r}^i = \mathbf{o} + t_\mathbf{r}^i \mathbf{v}$ ($t_n \le t_\mathbf{r}^i \le t_f$, $t_n$ and $t_f$ denote the near and far bounds) along the ray $\mathbf{r}$ are sampled, based on the SDF value $s_\mathbf{r}^i$ of each point $\mathbf{p}_\mathbf{r}^i$, the density value $\sigma_\mathbf{r}^i$ can be obtained as:
\begin{equation}
	\sigma_\beta(s)=\left\{\begin{array}{ll}
		\frac{1}{2 \beta} \exp \left(\frac{s}{\beta}\right) & s \leq 0 \\
		\frac{1}{\beta}\left(1-\frac{1}{2} \exp \left(-\frac{s}{\beta}\right)\right) & s>0
	\end{array},\right.
	\label{eq1}
\end{equation}
where $\beta$ is a learnable parameter. Subsequently, the density $\sigma_\beta$ is used to derive the color field through the volume rendering technique \cite{Mildenhall2020}. The color $\hat{C}(\mathbf{r})$ for the ray $\mathbf{r}$ is computed by:
\begin{equation}
	\hat{C}(\mathbf{r})=\sum_{i=1}^P T_{\mathbf{r}}^i \alpha_{\mathbf{r}}^i \hat{\mathbf{c}}_{\mathbf{r}}^i, \quad T_{\mathbf{r}}^i=\prod_{j=1}^{i-1}\left(1-\alpha_{\mathbf{r}}^j\right),
	\label{eq2}
\end{equation}
where $T_{\mathbf{r}}^i$ and $\alpha_{\mathbf{r}}^i=1-\exp \left(-\sigma_{\mathbf{r}}^i \delta_{\mathbf{r}}^i\right)$ denote the transmittance and alpha value of the $i$-$th$ point along ray $\mathbf{r}$, respectively, and $\delta_{\mathbf{r}}^i$ is the distance between neighboring sample points. The  normal $\hat{N}(\mathbf{r})$ of the surface intersecting the ray $\mathbf{r}$ can be obtained as follows:
\begin{equation}
	\hat{N}(\mathbf{r})=\sum_{i=1}^P T_{\mathbf{r}}^i \alpha_{\mathbf{r}}^i \hat{\mathbf{n}}_{\mathbf{r}}^i.
	\label{eq3}
\end{equation}

\subsubsection{Multi-res Hash Encoding}
Multi-res hash encoding \cite{Mueller2022} presents an efficient technique for encoding input coordinates into a high-dimensional space, thereby facilitating the learning of high-frequency details. To enhance the reconstruction ability of PSDF for fine-grained details, the multi-res hash encoding is introduced to encode samples $\mathcal{T}$ obtained by the error-bounded sampling strategy and the proposed depth prior-assisted sampling as shown in Fig. \ref{framework}.

Multi-res hash encoding establishes multi-res hash grids, where each corner of a grid cell is mapped to a corresponding hash entry, serving as a container for storing encoding features. To elaborate, given multi-resolution hash grids $\left \{ \Phi_\theta^l\right \} _{l=1}^L$ with the resolution $R_l: = \left \lfloor R_{\textrm{min}}b^l \right \rfloor $ ($b:=\exp(\frac{\ln R_{\textrm{max}} - \ln R_{\textrm{min}}}{L-1} )$, $R_{\textrm{max}}$ and $R_{\textrm{min}}$ denote the coarsest and finest resolution), we can get encoded features $h^l(\mathbf{p};\Phi_\theta^l)$ of a query position at each level's hash grids by interpolation and concatenated them together to form a feature vector $h(\mathbf{p})$:
\begin{equation}
	h(\mathbf{p}) = \left \{h^1(\mathbf{p};\Phi_\theta^1), \dots , h^L(\mathbf{p};\Phi_\theta^L)  \right \}.
	\label{eq4}
\end{equation}

\subsection{External Geometric Prior-guided Learning}
\label{sec:external_geometric_prior-guided_learning}
\subsubsection{Visibility-aware Feature Consistency Loss}

The visibility-aware feature consistency loss is used to enhance the geometric consistency of the recovered geometry by optimizing learned surface intersecting points. Unlike existing methods, our approach leverages deep features and visibility maps from a learning-based MVS, overcoming issues of limited representative ability of hand-crafted metrics and erroneous occlusion constraints. This leads to more precise photoconsistency constraints and improved quality of the reconstructed surface.

To achieve this, we first find the location $\hat{\mathbf{p}}$ of surface $\mathcal{S}$ from SDFs. Recalling that $P$ points $\mathbf{p}_\mathbf{r}^i =  \mathbf{o} + t_\mathbf{r}^i \mathbf{v} (i=1, \dots, P)$ are sampled along the ray $\mathbf{r}$ by error-bounded sampling strategy \cite{Yariv2021} to render a pixel $x$ during training.
The $\mathcal{S}$ is located between the first two points $\mathbf{p}_\mathbf{r}^{i}$ and $\mathbf{p}_\mathbf{r}^{i+1}$ whose sign of the SDF values are different, namely, the location $\hat{\mathbf{p}}$ of $\mathcal{S}$ can be approximated by:
\begin{equation}	
	\begin{aligned}
		\hat{\mathbf{p}} = \mathbf{o} + \hat{t}  \mathbf{v}  , \quad
		\hat{t} = \textrm{argmin}\left \{ t | t \in \hat{\mathcal{T}}^*\right \}, \\
		\hat{\mathcal{T}}^* = \left \{ t | t = \frac{f_\theta(\mathbf{p}_\mathbf{r}^i)t_\mathbf{r}^{i+1} - f_\theta(\mathbf{p}_\mathbf{r}^{i+1})t_\mathbf{r}^i}{f_\theta(\mathbf{p}_\mathbf{r}^i) - f_\theta(\mathbf{p}_\mathbf{r}^{i+1})}, t \in \mathcal{T} \right \}.
	\end{aligned}
	\label{eq5}
\end{equation}

Based on the points $\hat{\mathbf{p}}$ of pixel $x$ in the current target (reference) view $I_r$ on the surface $\mathcal{S}$, we can obtain the corresponding pixel $x_m$ in the $m$-$th$ source view $I_m$ by the plane-induced homography $\mathbf{H}_m$\cite{Hartley2003}:
\begin{equation}
	\mathbf{H}_m = \mathbf{K}_m (\mathbf{R}_m\mathbf{R}_r^T - \frac{\mathbf{R}_m(\mathbf{R}_m^T\mathbf{t}_{m} - \mathbf{R}_r^T\mathbf{t}_{r})\mathbf{n}^T}{d})\mathbf{K}_r^{-1},
	\label{eq6}
\end{equation}
where $\mathbf{K}_r$, $\mathbf{R}_r$, $\mathbf{t}_{r}$ and $\mathbf{K}_m$, $\mathbf{R}_m$, $\mathbf{t}_{m}$ denote the intrinsic parameters, rotation and translation of reference view and $m$-$th$ source view, respectively. 

In this case, features of pixel $x$ and features of the corresponding pixel $x_m$ can be extracted from the image feature extraction network of a pretrained MVS network.
To enhance the representation ability, instead of using pixel-wise features for consistency constrain, the features $F_r$ of the $Q \times Q$ patch $q_r$ with pixel $x$ centered and the corresponding features $F_m$ of patch $q_m$ are extracted. The photometric consistency of $F_r$ and $F_m$ is measured by the cosine similarity:
\begin{equation}
	\mathcal{C}_m = \frac{F_r \cdot F_m}{\left \| F_r \right \|_2 \left \| F_m \right \|_2},
	\label{eq7}
\end{equation}
where  $\left \| \cdot  \right \| _2$ denotes the 2-norm.

To handle the visibility problem, visibility maps predicted by the pretrained MVS network are further introduced, and the visibility-aware feature consistency loss is given by
\begin{equation}
	\mathcal{L}_{\textrm{vfc}} = \frac{ {\textstyle \sum_{n_p=1}^{N_p}}{\textstyle \sum_{m=1}^{M}} V_m {\textstyle \sum_{n_q=1}^{N_q}} (1 - \mathcal{C}_m )}{N_pMN_q},
	\label{eq8}
\end{equation}
where $N_p$ and $M$ denote the number of pixels in the minibatch and the number of source views. $N_q$ denotes the number of pixels in the patch $q_r$.

\subsubsection{Depth Prior-assisted Sampling}
The sampling strategies employed in NISR \cite{Wang2021, Yariv2021} are designed to generate samples that closely approximate surface intersecting points $\hat{\mathbf{p}}$. The depth generated through MVS contains valuable cues indicating the probable positions of these surface intersecting points.
To improve the efficiency of locating these points, depth prior-assisted sampling is introduced, which utilizes depth priors from MVS to guide the generation of informed samples. 
It is combined with error-bounded sampling \cite{Yariv2021} to ensure accurate and effective sample generation.

The process begins by filtering the depth produced by a pretrained MVS Network using photometric consistency and geometric consistency criteria, following the approach of previous learning-based MVS methods \cite{Yao2018}. This filtering helps eliminate unreliable depth estimates. The resulting reliable depth value, denoted as $d_{\textrm{mvs}}$, for a given pixel $x$, is projected into the 3D space of NISR. This projection yields a 3D point $\mathbf{p}_{\textrm{mvs}}$, which is subsequently transformed into a distance $t_{\textrm{mvs}}$ from the camera center $\mathbf{o}$ along the ray $\mathbf{v}$ using the formula $t_{\textrm{mvs}} = \frac{\mathbf{p}_{\textrm{mvs}} - \mathbf{o}}{\mathbf{v}}$.

Drawing inspiration from the coarse-to-fine strategy used in CasMVSNet \cite{Gu2020}, we generate $P_{\textrm{mvs}}$ samples that potentially contain points very close to or located at the surface uniformly from the range $[t_{\textrm{mvs}} - \frac{H_tP_{\textrm{mvs}}}{2}, t_{\textrm{mvs}} + \frac{H_tP_{\textrm{mvs}}}{2}]$. Here, $P_{\textrm{mvs}}$ represents the number of samples generated, $H_t = \frac{t_f - t_n}{P_{t}}$ denotes the hypothesis interval between the $i$-$th$ and ($i$+1)-$th$ samples, and $P_{t}$ is a predefined scalar used to control the hypothesis interval $H_t$.
The $P_{\textrm{mvs}}$ samples generated using the above steps are then combined with the $P$ samples generated by the error-bounded sampling strategy to form the final set of samples $\mathcal{T}$.

Note that VolSDF \cite{Yariv2021} not only obtains some samples by inverse transform sampling based on the candidate samples $\mathcal{T}^*$ and $\beta_+$ that are estimated by the error-bounded sampling strategy but also samples some extra samples by randomly sampling from $\mathcal{T}^*$. For the pixels without depth priors, we follow the mechanism of VolSDF to get the final samples. For pixels with depth priors, we do not perform random sampling, but instead generate some samples based on the depth prior-assisted sampling as mentioned earlier to obtain the final sample.

\subsection{Internal Prior-guided Importance Rendering}
\label{sec:internal_prior-guided_importance_rendering}
The inherent prior embedded in NISR is further leveraged to mitigate the bias arising from disparities between surface rendering and volume rendering. While it is ideal for the learned colors of surface points to match the RGB image, challenges arise due to biases originating from volume rendering integrals and densely distributed near-surface points-based rendering, rendering the ideal situation hard to attain.
Both the points obtained by the zero-level set of SDFs and the depth acquired through volume rendering integrals provide indications of surface intersecting points. Be aware of this, we exploit these two cues to facilitate importance rendering, thereby steering the model's focus towards learning accurate colors for surface intersection points.

To achieve this, we deduce the positions $\hat{\mathbf{p}}$ at the zero-level set of SDFs and $\tilde{\mathbf{p}}$ at the rendered depth using Eq. (\ref{eq5}) and Eq. (\ref{eq9}), respectively.
\begin{equation}
	\hat{D}(\mathbf{r})=\sum_{i=1}^P T_{\mathbf{r}}^i \alpha_{\mathbf{r}}^i t_{\mathbf{r}}^i, \quad
	\tilde{\mathbf{p}} = \mathbf{o} + \hat{D}  \mathbf{v}.
	\label{eq9}
\end{equation}

The informed points $\hat{p}$ and $\tilde{p}$ form a new set of samples $\mathcal{T}_{\textrm{IR}}$, which is augmented by $Q$ samples uniformly extracted from the range $[t_n, t_f]$ along the ray $\mathbf{v}$. This new set of samples, $\mathcal{T}_{\textrm{IR}}$, is employed in a forward and backward pass for internal prior-guided importance rendering during the training process.
This enables the model's focus on unbiased color learning for nearby surface points.

In this stage, only the color loss is used:
\begin{equation}
	\mathcal{L}_{\textrm{rgb}}^{\textrm{IR}}=\sum_{\mathbf{r} \in \mathcal{R}}\|\hat{C}^{\textrm{IR}}(\mathbf{r})-C(\mathbf{r})\|_1,
	\label{eq10}
\end{equation}
where the $\mathcal{R}$ denotes the set of rays in the minibatch, $C(\mathbf{r})$ denotes the ground-truth pixel color, and $\hat{C}^{\textrm{IR}}(\mathbf{r})$ denotes the rendered color resulting from the internal prior-guided importance rendering.

\subsection{Optimization}
\label{sec:optimization}
In the non-importance rendering stage, except for the introduced visibility-aware feature consistency loss $\mathcal{L}_{\textrm{vfc}}$, the normal consistency loss $\mathcal{L}_{\textrm{normal}}$ \cite{Yu2022} for constraining low-textured areas, the geometry bias loss $\mathcal{L}_{\textrm{bias}}$ \cite{Chen2023} for regularizing the biased reconstruction, and the smooth loss $\mathcal{L}_{\textrm{smooth}}$ \cite{Oechsle2021} for encouraging the smoothness of the recovered surfaces are used. Moreover, the
color loss $\mathcal{L}_{\textrm{rgb}}$ and Eikonal Loss $\mathcal{L}_{\textrm{eikonal }}$ are also used following the common practice for optimizing the scene representation and regularizing SDF values, respectively.

The normal consistency  is measured by angular and L1 losses \cite{Yu2022}:
\begin{equation}
	\mathcal{L}_{\textrm{normal}}=\sum_{\mathbf{r} \in \mathcal{R}}\|\hat{N}(\mathbf{r})-\bar{N}(\mathbf{r})\|_1+\|1-\hat{N}(\mathbf{r})^{\top} \bar{N}(\mathbf{r})\|_1,
	\label{eq11}
\end{equation}
where $\hat{N}$ denotes normal rendered from the geometry network, $\bar{N}$ denotes the monocular normal predicted by the pretained Omnidata model \cite{Eftekhar2021}.

The geometry bias loss $\mathcal{L}_{\textrm{bias}}$ \cite{Chen2023} is given by
\begin{equation}
	\mathcal{L}_{\textrm{bias}} = \frac{1}{|\mathcal{S}|} \sum_{\mathbf{\hat{p}}\in \mathcal{S}}|f_\theta(\mathbf{\hat{p}})|,
	\label{eq12}
\end{equation}
where $\hat{p}$ denotes the points at the zero-level set of the learned SDFs,  $\mathcal{S}$ denotes the set points $\hat{p}$ in a minibatch, $|\mathcal{S}|$ denotes the number of points $\hat{p}$, $|f_\theta(\mathbf{\hat{p}})|$ denotes the absolute value of $f_\theta(\mathbf{\hat{p}})$.

The  smooth loss $\mathcal{L}_{\textrm{smooth}}$ \cite{Oechsle2021} encourages the normal of a  point $\mathbf{p}$ and its neighboring points $\mathbf{p} + \varepsilon$ to be similar:
\begin{equation}
	\mathcal{L}_{\textrm{smooth}} = \sum_{\mathbf{r} \in \mathcal{X}}\|\hat{\mathbf{n}}_{\mathbf{r}}(\mathbf{p})-\hat{\mathbf{n}}_{\mathbf{r}}(\mathbf{p}+ \varepsilon)\|_2,
	\label{eq13}
\end{equation}
where $\mathcal{X}$ are a set of uniformly sampled points together with near-surface points, the normal at $\mathbf{p}$ is computed by
\begin{equation}
	\hat{\mathbf{n}}_{\mathbf{r}}(\mathbf{p})= \frac{\bigtriangledown f_\theta(\mathbf{p})}{\left \|  f_\theta(\mathbf{p})\right \|_2 }. 
	\label{eq14}
\end{equation}

The color loss $\mathcal{L}_{\textrm{rgb}}$ and Eikonal Loss $\mathcal{L}_{\textrm{eikonal}}$ are given as follows,
\begin{equation}
	\mathcal{L}_{\textrm{rgb}}=\sum_{\mathbf{r} \in \mathcal{R}}\|\hat{C}(\mathbf{r})-C(\mathbf{r})\|_1
	\label{eq15},
\end{equation}

\begin{equation}
	\mathcal{L}_{\textrm{eikonal}}=\sum_{\mathbf{p} \in \mathcal{X}}\left(\left\|\nabla f_\theta(\mathbf{p})\right\|_2-1\right)^2.
	\label{eq16}
\end{equation}

In summary, the total loss used for the entire training process is as follows:
\begin{equation}
	\begin{aligned}
		\mathcal{L} &= \mathcal{L}_{\textrm{rgb}} + \gamma_1 \mathcal{L}_{\textrm{eikonal }} + \gamma_2 \mathcal{L}_{\textrm{vfc}} + \gamma_3 \mathcal{L}_{\textrm{normal}} \\
		& + \gamma_4 \mathcal{L}_{\textrm{bias}} + \gamma_5 \mathcal{L}_{\textrm{smooth}} + \gamma_6 \mathcal{L}_{\textrm{rgb}}^{\textrm{IR}}
	\end{aligned}
	\label{eq17}
\end{equation}
where $\gamma_i$ ($i=1, \dots, 6$) are the weight coefficients.

\section{Experiments}
\subsection{Datasets and Evaluation Metrics}
Experiments are conducted on Tanks and Temples \cite{Knapitsch2017} and DTU \cite{Aanaes2016} datasets.
Tanks and Temples \cite{Knapitsch2017} includes large-scale indoor and outdoor scenes with 151 to 1107 views captured under real-world unconstrained environment. It is split into training  subset, intermediate subset and advanced subset with 7 scenes, 8 scenes and 6 scenes, respectively. 
DTU contains various object-centric scenes with 49 or 64 views captured under controllable environment. Following previous works \cite{Yariv2020}, 15 scenes are selected for evaluation on DTU.

For Tanks and Temples dataset, except for the training set, ground-truths
are unavailable for the public. Reconstruction results need to be uploaded to the official website to obtain the evaluation results, so we follow the official to use F$_1$ score as the metric on this dataset which measures the precision and recall of the vertices of reconstructed meshes based on the GT point clouds.
The Python script\footnote{https://github.com/isl-org/TanksAndTemples/tree/master/pyt-\\hon$\_$toolbox/evaluation} provided by the official is used to compute these evaluation metrics on the training subset.
While for the intermediate subset and advanced subset, we submit the reconstructed results to the evaluation server\footnote{https://www.tanksandtemples.org/} to compute these evaluation metrics.
The results of COLMAP \cite{Schoenberger2016}, MVS Prior, VolSDF \cite{Yariv2021}, MonoSDF \cite{Yu2022} and PSDF are shown on the Leaderboard, which is named COLMAP$\_$Mesh$\_$unofficial, MVS$\_$Prior, VolSDF$\_$unofficial, MonoSDF$\_$unofficial and PSDF on the Leaderboard.

For DTU dataset, we follow the official evaluation protocol and report the reconstruction quality with Chamfer Distance which is the mean of Accuracy and Completeness. Similar to previous methods \cite{Yu2022}, the Python script\footnote{https://github.com/jzhangbs/DTUeval-python} is used to compute the evaluation metric for efficiency.

\subsection{Implementation Details and Experimental Settings}
\subsubsection{PSDF}
The PSDF is implemented by PyTorch \cite{Paszke2019} with Adam \cite{Kingma2014} as the optimizer. During training, the learning rate is set to 1e-2 for hash grids and 5e-4 for the rest of the components, and it decays exponentially. The training process consists of 200$K$ iterations on the Tanks and Temples dataset and 100$K$ iterations on the DTU dataset, with a minibatch size of 1024 pixels.
On each ray $\mathbf{r}$, 64 samples are sampled by the error-bounded sampling \cite{Yariv2021}, 32 samples are sampled by the depth-assisted sampling.
For the outdoor scenes in Tanks and Temples, we use the contraction \cite{Barron2022} to model the complex background, namely, in addition to the 96 points sampled by the error-bounded sampling \cite{Yariv2021} and depth-assisted sampling, we sample 32 points outside the defined sphere.
For the multi-res hash encoding, $R_{\textrm{min}} =2^4$, $R_{\textrm{max}} = 2^{11}$, $L=16$, the maximum number of hash entries per level and feature channel per level are set to 2$^{22}$ and 2.
The geometry network and color network are modeled by 2 layer MLPs with a hidden size 256, and the geometry network is initialized by geometric initialization \cite{Atzmon2020}.
Positional encoding \cite{Mildenhall2020} is applied to the query position $\mathbf{p}$ with 6 frequencies and the view direction $\mathbf{v}$ with 4 frequencies.
To adapt PSDF to variable lighting of different views, appearance embedding \cite{MartinBrualla2021} are used in Tanks and Temples.
 
The triangular mesh is extracted form the geometry network using Marching Cube \cite{William1987}, and space resolution is set to  $2048^3$ for Tanks and Temples and $512^3$ for DTU.
All experiments are done on a GeForce RTX 2080Ti GPU.

The prior MVS network, trained on the BlendedMVS dataset \cite{Yao2020} with 10 epochs, mainly adopts the architecture of CasMVSNet \cite{Gu2020} and incorporates visibility-aware cost volume construction \cite{Xu2022b} and uncertainty-guided training \cite{Su2022} for visibility estimation and uncertainty estimation.
The resolution of the image for training is 640 $\times$ 512, and the number of views is set to 7. The number of depth hypothesis for each staged used for training and test are 32, 16 and 8, respectively. The learning rate is set to 0.01 and decreased by half at $6$-$th$, $8$-$th$ and $10$-$th$ epochs. And the robust training strategy \cite{Wang2021a} is used for training. The loss weights for each stage are 0.5, 1 and 2, respectively.
In the process of generating feature, visibility and depth priors, the number of views are set to 11 on both and Tanks and Temples \cite{Knapitsch2017} and  DTU \cite{Aanaes2016} datasets, the resolution of the image used is $1024 \times 576$ for Tanks and Temples and  $1600 \times 1152$ for DTU. 
The photometric consistency based on the estimated uncertainty map of 3 stages and geometric consistency \cite{Yao2018} are used to filter and fuse the depth maps to a unified 3D point cloud.
More specifically, the uncertainty map is converted to the certainty map for filtering unreliable estimates, and filtering thresholds for the 3 stages' uncertainty maps are 0.6, 0.6 and 0.6, respectively. When performing geometric consistency, the thresholds for reprojected coordinate error and the reprojected depth error are 1 and 0.01, and the threshold for the number of view consistent is 5. 

\subsubsection{Compared Methods}
For  Tanks and Temples \cite{Knapitsch2017} datasets, except for Neuralangelo \cite{Li2023} which provides the results of some methods on six out of the seven scenarios of training subset, there are no public results of NISR on this dataset available. 
Since NISR methods require per-scene optimization, it is not feasible to use many methods to retrain on this dataset for comparison which will take a lot of time.
For this reason, we additionally train two typical NISR methods which are closely related to PSDF on this dataset for comparison, i.e., VolSDF \cite{Yariv2021} and MonoSDF \cite{Yu2022} (Multi-res hash encoding-based version). Moreover, we also give the results of classic MVS-based methods (COLMAP \cite{Schoenberger2016} and MVS Prior) on this dataset. 
Note that the MVS Prior denotes the priors used in the depth-assisted sampling strategy, namely, the reliable depths used are fused into a unified point cloud and transformed into a surface mesh.

\noindent \textbf{VolSDF} We follow the official implementation of VolSDF\footnote{https://github.com/lioryariv/volsdf} \cite{Yariv2021} and use the default settings to conduct the experiments on the Tanks and Temple dataset. Specifically, for indoor scenes, we adopt their settings on the DTU dataset, while for outdoor scenes, we adopted their settings on BlendedMVS dataset. The training process consists of 200$K$ iterations on all scenes of the Tanks and Temple dataset, with a minibatch size of 1024 pixels.

\noindent \textbf{MonoSDF} We follow the official implementation of MonoSDF\footnote{https://github.com/autonomousvision/monosdf} \cite{Yu2022} to perform the experiments on the Tanks and Temple dataset, where the multi-res hash encoding-based version is used. The setting for the MonoSDF is the same as PSDF, namely, the settings of the learning rate,  multi-res hash grid, appearance embeddings, and so on are the same as PSDF. Note that the MonoSDF does not provide the model for handling outdoor scenes, so we use the contraction \cite{Barron2022}  to model the background of the outdoor scenes in Tanks and Temples, which is also the same as that of PSDF. The training process consists of 200$K$ iterations on all scenes of the Tanks and Temple dataset, with a minibatch size of 1024 pixels.

\noindent \textbf{Classic MVS-based Methods} For the classic MVS-based methods, COLMAP \cite{Schoenberger2016} and MVS Prior, we use the sPSR \cite{Kazhdan2013} to recover the surface mesh from the point clouds reconstructed by them. 
To align with the resolution used in the Marching Cube \cite{William1987} for NISR methods, the depth is set to 12 and 9 on the Tanks and Temples and DTU datasets, respectively. The trim is set to 7 on these two datasets. Note that for the results of COLMAP \cite{Schoenberger2016} on the DTU, we adopt the results presented in the previous works \cite{Wang2021}.

Since that when MVS reconstructs outdoor scenes, the background and sky are reconstructed together, resulting in a particularly large scene range for reconstructed point clouds. That is to say, there are many points far away from the object of interest. In this case, naive applying the sPSR \cite{Kazhdan2013} for surface reconstruction on the original reconstructed point clouds by COLMAP \cite{Schoenberger2016}  will result in poor reconstruction of the region of interest.
To this end, we manually remove the points that are far from the object of interest before applying sPSR. 
The mean F$_1$ scores of the outdoor scenes reconstructed by COLMAP \cite{Schoenberger2016} in the training set of Tanks and Temples datasets with and without manual cropping are 1.03\% and 11.83\%, respectively.
It can be seen that COLMAP \cite{Schoenberger2016} can not obtain reasonable results on outdoor scenes without manual cropping. 
For the results of outdoor scenes listed in Table \ref{tab:tanks-training} and Table \ref{tab:tanks-test}, the manual cropping are applied.
Since the MVS Prior can obtain reasonable results on outdoor scenes without any manual cropping, we do not apply this operation on it.

\begin{table*}[htbp]
	\caption{Quantitative results on the training  subset of Tanks and Temples with F$_1$ scores (\%) (higher is better). $^\ast $ denotes the results adopted from Neuralangelo \cite{Li2023}. 
		Methods are separated into classic MVS-based methods and NISR methods (from top to bottom).
	}
	\label{tab:tanks-training}
	\centering
	\small
	\begin{tabular}{l|ccccccc|c}
		\hline
		Method & Church & Meetingroom & Barn &Caterpillar & Courthouse & Ignatius & Truck  & Mean\\ \hline
		COLMAP \cite{Schoenberger2016} & 39.90 & 5.25& 34.20 & 3.23 & 23.60 & 21.05 & 22.43 & 21.38 \\
		MVS Prior & \underline{48.82} & \underline{36.8} & 31.64 & 10.15 & \underline{37.61} & 10.15 & 22.49 & 28.23 \\ \hline
		VolSDF \cite{Yariv2021} & 3.84 & 3.67 & 12.00 & 13.18 & 6.50 & 17.77 & 25.15& 11.73\\
		NeuralWarp$^\ast $ \cite{Darmon2022} & - & 8 & 22 & 18 & 8 & 2 & 35 & 15 \\
		MonoSDF \cite{Yu2022} & 29.04 & 33.92 & 35.21 & 19.34 & 22.10  & 31.61 &41.03 &30.32\\
		Geo-Neus$^\ast $ \cite{Fu2022} & - & 20 & 33 & 26 & 12 & 72 & 45&35\\
		Neus$^\ast $ \cite{Wang2021} & - & 24 & 29 & 29 & 17 & \underline{83} &45&38\\
		Neuralangelo \cite{Li2023} & - & 32 & \textbf{70} & \underline{36} & 28 & \textbf{89} &\underline{48}&\underline{50}\\ 
		PSDF & \textbf{54.80} & \textbf{47.05} & \underline{62.27} & \textbf{38.51} & \textbf{41.65} & 78.80 & \textbf{52.65} & \textbf{53.68} \\
		\hline
	\end{tabular}
\end{table*}

\begin{table*}[htbp]
	\caption{Quantitative results on Intermediate and Advanced subsets of Tanks and Temples with F$_1$ scores (\%) (higher is better). 
		Methods are separated into classic MVS-based methods and NISR methods (from top to bottom).
	}
	\label{tab:tanks-test}
	\centering
	\small
	\begin{tabular}{l|p{0.47cm}<{\centering}p{0.47cm}<{\centering}p{0.47cm}<{\centering}p{0.47cm}<{\centering}p{0.47cm}<{\centering}p{0.47cm}<{\centering}p{0.47cm}<{\centering}p{0.45cm}<{\centering}p{0.6cm}<{\centering}|p{0.47cm}<{\centering}p{0.47cm}<{\centering}p{0.64cm}<{\centering}p{0.72cm}<{\centering}p{0.57cm}<{\centering}p{0.7cm}<{\centering}p{0.6cm}<{\centering}}
		\hline
		\multirow{2}{*}{Method}
		& \multicolumn{9}{c|}{Intermediate}&\multicolumn{7}{c}{Advanced}\\
		& Fam. & Franc. & Horse & Light. & M60 & Pan. & Play. & Train & Mean & Audi. & Ballr. & Courtr. & Muse. & Palace & Temple & Mean \\ \hline
		COLMAP \cite{Schoenberger2016} & 22.35 & 2.39 & 9.14& 42.28& 11.95 &21.87  & 29.04 &4.67  &17.96 &12.98 &12.30 & 25.16 & 34.72 & 9.24 &21.94 &19.39  \\
		MVS Prior & 10.49 & 1.88 & 0.70 & \underline{44.6} & \underline{34.45} & \underline{35.82} & 14.15 & 17.82 & 19.99 & \underline{16.95} & \underline{31.40} & \underline{30.02} & \underline{42.53} & \underline{14.63} & 11.71 & \underline{24.54} \\
		\hline
		VolSDF \cite{Yariv2021} & 21.94 & 8.75 & 21.82 & 16.46 & 14.27 & 21.95 & 1.90 & 10.24 & 14.67 & 3.16 & 11.61 & 7.71 & 4.41 & 2.40 & 4.44 & 5.62 \\
		MonoSDF \cite{Yu2022} &\underline{41.94} & \underline{20.37} & \underline{30.28} & 31.01 & 21.79 & 20.81 & \underline{44.67} & \underline{18.29} & \underline{28.65} & 10.97 & 29.30 & 21.58 & 21.71 & 7.04 & \underline{17.64 }& 18.04 \\
		PSDF & \textbf{67.85} & \textbf{34.43} & \textbf{49.30} & \textbf{51.02} & \textbf{38.37} & \textbf{51.46} & \textbf{47.26} & \textbf{38.82} & \textbf{47.31} & \textbf{26.56}& \textbf{42.90}& \textbf{36.88}&\textbf{45.77} &\textbf{16.36} &\textbf{34.43} & \textbf{33.82} \\
		\hline
	\end{tabular}
\end{table*}

\subsubsection{Data Pre-processing}
Adhering to the script\footnote{https://github.com/lioryariv/volsdf/tree/main/data/preprocess} provided by VolSDF \cite{Yariv2021}, we use the camera parameters processed by MVSNet \cite{Yao2018} for both DTU and Tanks and Temples datasets to shift the coordinate system, locating the object at the origin. 
When dealing with indoor scenes, we designate a sphere with a radius of 3, while for outdoor scenes, a sphere with a radius of 1 is employed. 
Note that due to VolSDF \cite{Yariv2021} using the parametrization of NeRF++ \cite{Zhang2020} to model the background, instead of using the contraction \cite{Barron2022}, the sphere's radius remains consistent at 3, irrespective of whether the scenes are indoor or outdoor.

To adapt the input resolution of the prior MVS network used, the resolution of images on the Tanks and Temples \cite{Knapitsch2017}  and DTU \cite{Aanaes2016} is cropped and resized to $1024 \times 576$ and $1600 \times 1152$, respectively.

The image of Tanks and Temples with resolution $1024 \times 576$ and the image of DTU with resolution $1600 \times 1152$ are input to the pretrained Omnidata model \cite{Eftekhar2021} to obtain monocular normal priors.

\subsection{Benchmark Performance}

\begin{figure*}[htb]
	\setlength{\abovecaptionskip}{1.5mm}
	\setlength{\belowcaptionskip}{0cm}
	\center
	\includegraphics[width=0.95\textwidth]{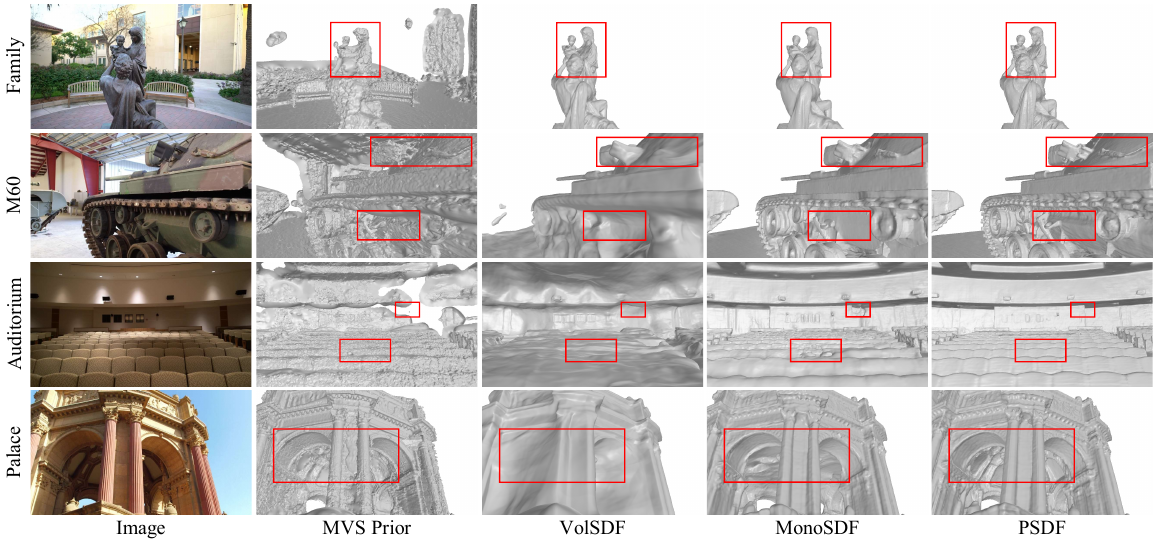}
	\caption{Qualitative comparison of surface meshes reconstructed by various methods on the Tanks and Temples dataset.}
	\label{tnt_visualization}
\end{figure*}

\subsubsection{Results on Tanks and Temples}
As shown in Table \ref{tab:tanks-training}, PSDF achieves the best performance on the training subset, surpassing other methods significantly. Compared with VolSDF \cite{Yariv2021}, MonoSDF \cite{Yu2022}, Geo-Neus \cite{Fu2022}, and Neus \cite{Wang2021}, PSDF demonstrates impressive improvements of 357.63$\%$, 77.04$\%$, 53.37$\%$, and 41.26$\%$, respectively.
The quantitative results on the intermediate subset and advanced subset presented in Table \ref{tab:tanks-test} also show the superiority of the PSDF compared with that of VolSDF \cite{Yariv2021} and MonoSDF \cite{Yu2022}. Our method achieves the best performance in all scenes of these two subsets.
Fig. \ref{tnt_visualization} shows the qualitative comparison of various methods. As shown, MVS Prior exhibit a considerable degree of roughness attributed to the negative impact of noise prevalent within the point cloud; VolSDF \cite{Yariv2021} is capable of restoring only the fundamental structure of the scene, showcasing limitations in capturing finer details; the performance of MonoSDF \cite{Yariv2021}  is susceptible to the unfaithful areas within monocular cues and shows limited reconstruction quality; in contrast, PSDF overcomes these limitations and demonstrates an impressive level of fidelity in its reconstructions, outperforming the other methods in delivering high-quality results.

\begin{table*}[htbp]
	\caption{Quantitative results on DTU dataset by using the Chamfer Distance (lower is better).  Methods are separated into classic MVS-based methods and NISR (from top to bottom). Note that for a fair comparison with NISR methods, the meshes of MVS Prior use the masks provided by \cite{Yariv2020} to crop the background areas as a post-processing step, MVS Prior$^\ast $ gives the results without such a post-processing step.
	}
	\label{tab:dtu}
	\centering
	\small
	\begin{tabular}{l|p{0.44cm}<{\centering}p{0.44cm}<{\centering}p{0.44cm}<{\centering}p{0.44cm}<{\centering}p{0.44cm}<{\centering}p{0.44cm}<{\centering}p{0.44cm}<{\centering}p{0.44cm}<{\centering}p{0.44cm}<{\centering}p{0.44cm}<{\centering}p{0.44cm}<{\centering}p{0.44cm}<{\centering}p{0.44cm}<{\centering}p{0.44cm}<{\centering}p{0.44cm}<{\centering}|c}
		\hline
		Method & 24 & 37 & 40 & 55 & 63 & 65 & 69 & 83 & 97& 105 & 106 & 110 & 114 & 118 & 122 & Mean \\ \hline
		COLMAP \cite{Schoenberger2016} & 0.81 & 2.05 & 0.73 & 1.22  & 1.79 & 1.58 & 1.02 & 3.05 & 1.40 & 2.05 & 1.00 & 1.32 & 0.49 & 0.78 & 1.17 & 1.36 \\
		MVS Prior & 0.52 & 0.77 & 0.35 & 0.41 & 0.88 & 1.12 & 0.63 & 1.16 & 1.02 & 0.62 & 0.70  & 0.58 & 0.38 & 0.50 & 0.49 & 0.68 \\
		MVS Prior$^\ast $ & 0.57 & 0.87 & 0.37 & 0.42 & 0.89 & 1.00 & 0.63 & \textbf{0.66} & \textbf{0.75} & 0.83 & 0.74  & 0.59 & 0.37 & 0.50 & 0.48 & 0.65 \\ \hline
		MVSDF \cite{Zhang2021}  & 0.83 & 1.76 & 0.88 & 0.44 & 1.11 & 0.90 & 0.75 & 1.26 & 1.02 & 1.35 & 0.87 & 0.84 & 0.34 & 0.47 & 0.47 & 0.89 \\
		VolSDF \cite{Yariv2021} & 1.14  & 1.26 & 0.81 & 0.49 & 1.25 & 0.70 & 0.72 & 1.29 & 1.18 & 0.70 & 0.66 & 1.08 & 0.42 & 0.61 & 0.55 & 0.86 \\
		Neus \cite{Wang2021} & 1.00 & 1.37 & 0.93 & 0.43  & 1.10 & 0.65 & 0.57 & 1.48 & 1.09 & 0.83 & 0.52 & 1.20 & 0.35 & 0.49 & 0.54 & 0.84 \\
		MonoSDF \cite{Yu2022} & 0.66 & 0.88 & 0.43 & 0.40 & 0.87 & 0.78 & 0.81 & 1.23 & 1.18 & 0.66 & 0.66 & 0.96 & 0.41 & 0.57 & 0.51 & 0.73 \\
		RegSDF \cite{Zhang2022} & 0.60 & 1.41 & 0.64 & 0.43 & 1.34 & 0.62 & 0.60 & \underline{0.90} & 0.92 & 1.02 & 0.60 & 0.59 & \underline{0.30} & 0.41 & 0.39 & 0.72 \\
		NeuralWarp \cite{Darmon2022}  & 0.49 & 0.71 & 0.38 & 0.38 & \underline{0.79} & 0.81 & 0.82 & 1.20 & 1.06 & 0.68 & 0.66 & 0.74 & 0.41 & 0.63 & 0.51 & 0.68 \\	
		NeuDA  \cite{Cai2023} & 0.47 & 0.71 & 0.42 & \underline{0.36}  & 0.88 & 0.56 & 0.56 & 1.43 & 1.04 & 0.81 & 0.51 & 0.78 & 0.32 & 0.41 & 0.45 & 0.65 \\
		D-Neus \cite{Chen2023} & 0.44 & 0.79 & \underline{0.35} & 0.39  & 0.88 & 0.58 & 0.55 & 1.35 & 0.91 & 0.76 & \textbf{0.40} & 0.72 & 0.31 & \underline{0.39} & 0.39 & 0.61 \\
		Neuralangelo \cite{Li2023} & \underline{0.37} & 0.72 & \underline{0.35} & \textbf{0.35}  & 0.87 & \underline{0.54} & 0.53 & 1.29 & 0.97 &  0.73 & 0.47 & 0.74 & 0.32 & 0.41  & 0.43 & 0.61 \\	
		Zhang et al. \cite{Zhang2023} & 0.49 & 0.71 & 0.37 & 0.36 & 0.80 & 0.56 & 0.52 & 1.17 & 0.97 & 0.66 & 0.48 & 0.73 & 0.32 & 0.42  & 0.42 & 0.60 \\
		Geo-Neus \cite{Fu2022} & 0.38 & \textbf{0.54} & \textbf{0.34} & \underline{0.36}  & 0.80 & \textbf{0.45} & \textbf{0.41} & 1.03 & \underline{0.84} & \textbf{0.55} & \underline{0.46} & \textbf{0.47} & \textbf{0.29} & \textbf{0.36} & \textbf{0.35} & \textbf{0.51}\\
		PSDF & \textbf{0.36} & \underline{0.60} & \underline{0.35} & \underline{0.36} & \textbf{0.70} & 0.61  & \underline{0.49} & 1.11 & 0.89 & \underline{0.60} & 0.47 & \underline{0.57} & \underline{0.30} & 0.40  & \underline{0.37} & \underline{0.55} \\
		\hline
	\end{tabular}
\end{table*}

\begin{figure*}[h]
	\center
	\includegraphics[width=0.90\textwidth]{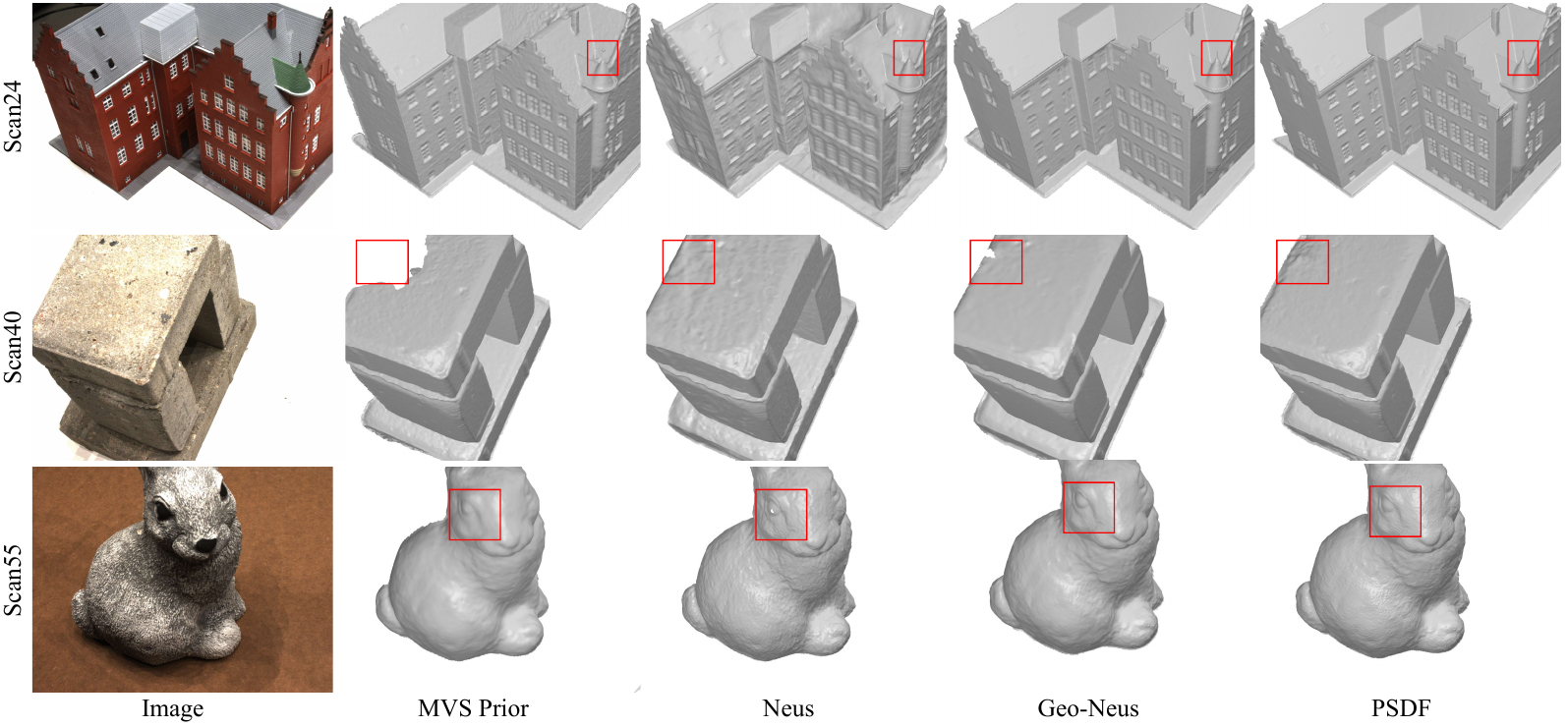}
	\caption{Qualitative comparison of surface meshes reconstructed by various methods on the DTU dataset.}
	\label{dtu_comparison}
\end{figure*}

\begin{figure}[htbp]
	\center
	\includegraphics[width=0.48\textwidth]{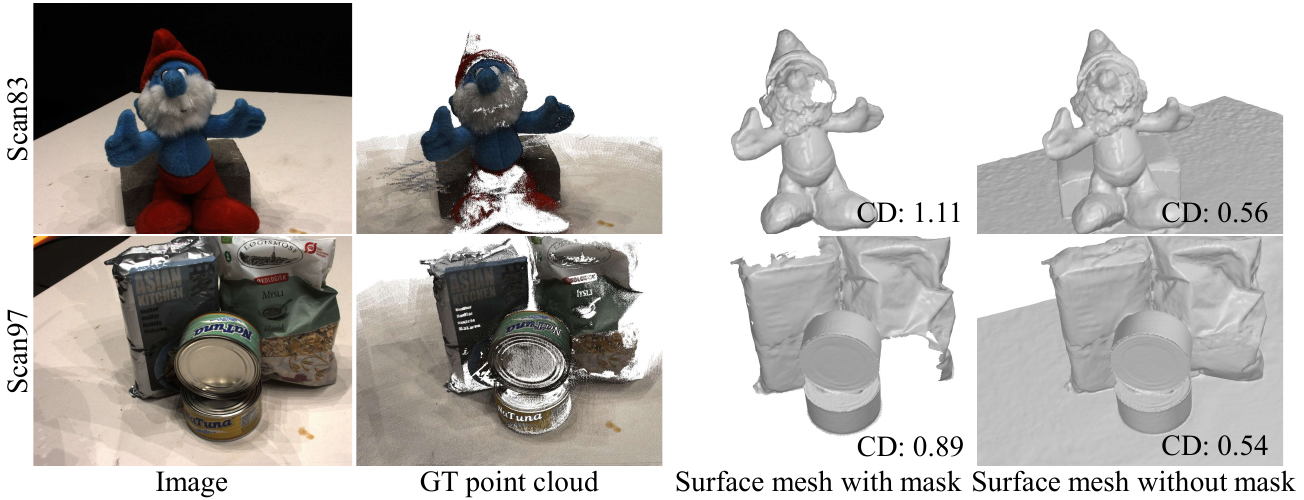}
	\caption{Visualization comparison of surface meshes reconstructed by PSDF with and without mask cropping on Scan83 and Scan97.}
	\label{dtu_mask}
\end{figure}

\subsubsection{Results on DTU}
Quantitative results of various surface mesh reconstruction methods on DTU dataset are given in Table \ref{tab:dtu}. As demonstrated, PSDF exhibits superior performance compared to both classic MVS-based methods and the majority of NISR methods including methods that also leverage priors, e.g., MVSDF \cite{Zhang2021}, MonoSDF \cite{Yu2022}, Zhang et al. \cite{Zhang2023}, D-Neus \cite{Chen2023}, Zhang et al. \cite{Zhang2023}. 
Additionally, PSDF achieves the second-best performance among existing NISR methods, with a 0.4 difference in Chamfer Distance when compared to the best-performing method, Geo-Neus \cite{Fu2022}.
This may be caused by the local optimization of the analytical gradient used in multi-res hash encoding as analyzed in Neuralangelo \cite{Li2023} and the difference in converting SDFs into density in VolSDF \cite{Yariv2021} and Neus \cite{Wang2021}. 
Notably, while Geo-Neus performs well on the simple scenes of DTU, it exhibits worse performance than PSDF on complex uncontrolled scenes in Tanks and Temples dataset. This is likely due to the adverse effects of hard-to-filter noise from the sparse point priors used in Geo-Neus. In contrast, PSDF demonstrates consistent performance even on complex uncontrolled scenes, highlighting its robustness and independence from the quality of priors.
Fig. \ref{dtu_comparison} illustrates the qualitative comparison of MVS Prior, Neus \cite{Wang2021}, Geo-Neus \cite{Fu2022} and PSDF on the DTU dataset. It can be seen that PSDF can recover delicate geometry.

As previous NISR methods all use masks provided by IDR \cite{Yariv2020} to crop the background areas as a post-processing step on the reconstructed meshes, we also apply this post-processing step on the PSDF for a fair comparison. We present the results with and without mask cropping as the post-processing step for MVS Prior in Table \ref{tab:dtu} which denote as ``MVS Prior'' and ``MVS Prior$^\ast $'', respectively. It can be seen from the results of ``MVS Prior'' and ``MVS Prior$^\ast $'', the meshes' quality is improved or decreased to varying degrees after the mask cropping.
Particularly noteworthy is the quality of meshes decreased a lot on Scan83 and Scan97 after the mask cropping, decreasing from 0.66 to 1.16 and 0.75 to 1.02, respectively. As shown in Fig \ref{dtu_mask}, as the masks provided by IDR \cite{Yariv2020} on these two scenes missing too many  foreground areas, many valid areas are missing after mask cropping. Without mask cropping, the Chamfer Distance (CD) of meshes reconstructed by PSDF on Scan83 and Scan97 are 0.56 and 0.54 which are better than that of MVS Prior$^\ast $. 
It can be seen that PSDF outperforms MVS Prior on all scenes, indicating that the quality of the prior depth used in the depth-assisted sampling strategy does not limit the performance of PSDF.

\begin{figure*}[htbp]
	\setlength{\abovecaptionskip}{1.5mm}
	\setlength{\belowcaptionskip}{0cm}
	\center
	\includegraphics[width=0.95\textwidth]{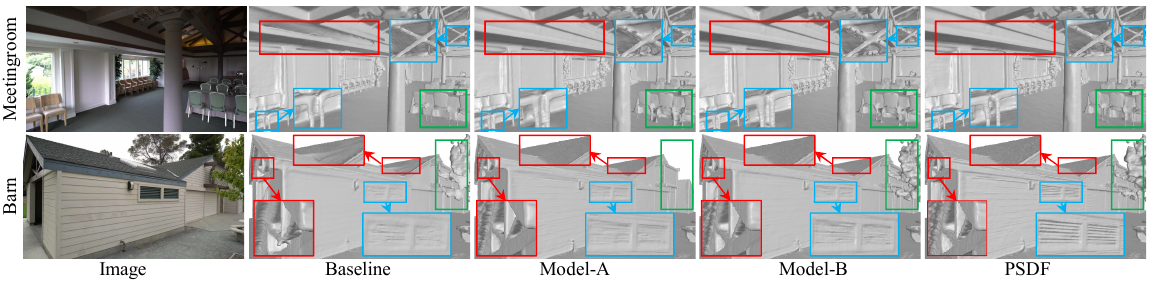}
	\caption{Visualization comparison of ablation study on ``Meetingroom" and ``Barn" of the Tanks and Temples dataset.}
	\label{Ablation-study}
\end{figure*}

\subsection{Ablation Study}
At first, to  validate the effectiveness of proposed components in the PSDF, the ablation study is conducted on the training subset of Tanks and Temples dataset. 
The Baseline model is configured as the full model PSDF does, but without Visibility-aware Feature Consistency (VFC) loss $\mathcal{L}_{\textrm{vfc}}$, Depth Prior-assisted Sampling (DPS), and Internal Prior-guided Importance Rendering (IPIR). 

Then, to further show the contribution of other components used in the Baseline model (i.e.,  Appearance Embedding (AE) \cite{MartinBrualla2021},  smooth loss $\mathcal{L}_{\textrm{smooth}}$ \cite{Oechsle2021}, normal consistency loss  $\mathcal{L}_{\textrm{normal}}$ \cite{Yu2022}, geometry bias loss $\mathcal{L}_{\textrm{bias}}$ \cite{Chen2023}), experiments are conducted based on Baseline model without these components and gradually enriched it through the incremental addition of these components. As NISR is a per-scene optimization method, performing experiments across all scenes within the training subset of Tanks and Temples entails a substantial time consumption. To this end, we opt for one indoor scene ``Church" and one outdoor scene ``Ignatius" of the training subset to conduct experiments which can also validate the effectiveness of each component. The mean precision, recall and $F_1$ score of the two scenes are reported.

\begin{table}[htbp]
	\caption{Ablation experimental results on the training subset of Tanks and Temples (higher is better).}
	\label{tab:Ablation-study}
	\centering
	\small
	\resizebox{\linewidth}{!}{
	\begin{tabular}{cccc|ccc}
		\hline
		Model & $\mathcal{L}_{\textrm{vfc}}$ & DPS & IPIR & Precision $\uparrow$ & Recall $\uparrow$ & $F_1$ $\uparrow$ \\
		\hline
		Baseline &  &  & & 26.33 & 35.44 & 29.43  \\
		Model-A & \checkmark & & & 46.31 & 58.20 & 50.72 \\
		Model-B & \checkmark & \checkmark & &46.20 & 59.52 & 51.49\\
		PSDF& \checkmark & \checkmark & \checkmark & \textbf{49.10} & \textbf{61.32}&\textbf{53.68} \\
		\hline
	\end{tabular}
    }
\end{table}

\subsubsection{Visibility-aware Feature Consistency Loss}
The experimental results without and with the $\mathcal{L}_{\textrm{vfc}}$ are shown in the first and second rows in Table \ref{tab:Ablation-study}, respectively. As illustrated, benefited from the powerful geometric constrain provided by the $\mathcal{L}_{\textrm{vfc}}$, the precision and recall of the surface mesh reconstructed by Model-A are greatly improved compared with that of Baseline. It can be seen from Fig. \ref{Ablation-study}, Model-A can reconstructed more geometrically consistent surfaces compared to Baseline, especially in some plane areas.

To further show the superiority of the proposed $\mathcal{L}_{\textrm{vfc}}$, it is  compared with the NCC-based photometric consistency loss $\mathcal{L}_{\textrm{ncc}}$ used in Geo-Neus \cite{Fu2022} and the multi-view feature consistency loss $\mathcal{L}_{\textrm{mfc}}$ used in D-Neus \cite{Chen2023} by combining them with the Baseline. 
As shown in Table \ref{tab:ph-loss}, Model-A shows better performance than that of Model-C and Model-D. Thanks to the discriminative features for robust similarity measure and visibility information for occlusion handling, the $\mathcal{L}_{\textrm{vfc}}$ can guide the optimization of geometry field better.

\begin{table}[htbp]
	\caption{Comparison results of different photometric consistency constraint on the training subset of Tanks and Temples (higher is better). }
	\label{tab:ph-loss}
	\centering
	\small
	\resizebox{\linewidth}{!}{
		\begin{tabular}{cccc|ccc}
			\hline
			Model & $\mathcal{L}_{\textrm{mfc}}$ & $\mathcal{L}_{\textrm{ncc}}$ & $\mathcal{L}_{\textrm{vfc}}$ & Precision $\uparrow$ & Recall $\uparrow$ & $F_1$ $\uparrow$ \\
			\hline
			Model-C & \checkmark &  &  & 36.12 & 49.94 &  40.71 \\
			Model-D & & \checkmark & & 43.04 &54.56 &47.28 \\
			Model-A &  & & \checkmark& \textbf{46.31} & \textbf{58.20} & \textbf{50.72} \\
			\hline
		\end{tabular}
	}
\end{table}

The features utilized in $\mathcal{L}_{\textrm{vfc}}$ are extracted from the feature pyramid network \cite{Lin2017} configured as CasMVSNet \cite{Gu2020} does but the output channel is set to 4 instead of 8 for improving computational efficiency. 
Using more advanced or complex network architectures (e.g., Transformer \cite{Darmon2022}) for feature extraction may further improve the performance of PSDF as it can improve the representative ability of the features.
In this paper, we do not explore the use of various features for $\mathcal{L}_{\textrm{vfc}}$ and leave it as future work.

\subsubsection{Depth Prior-assisted Sampling}
The comparison results presented in the second and third rows in Table \ref{tab:Ablation-study} show that the DPS contributes to improving the recall of the reconstructed surface. As illustrated in Fig. \ref{Ablation-study} of ``Barn", thanks to that DPS provides the samples generated by the depth prior as a supplement, Model-B can reconstruct the tree in the scene that is overlooked by Model-A.

\subsubsection{Internal Prior-guided Importance Rendering}
The presence of biased surface rendering, stemming from volume rendering, poses challenges to achieving accurate surface reconstruction. While the geometry bias loss $\mathcal{L}_{\textrm{bias}}$ and the visibility-aware feature consistency loss $\mathcal{L}_{\textrm{vfc}}$ alleviate this issue to a certain extent, the biased surface rendering still exists as the unchanged working mechanism of volume rendering. To make the network focus more on the color learning of surface intersecting points, the internal prior-guided importance rendering is introduced to further mitigate the biased surface rendering.
From the comparison results of the third and fourth rows of the Table \ref{tab:Ablation-study}, it can be seen that the internal prior-guided importance rendering can boost both precision and recall of recovered geometry. It can also be observed from Figure \ref{Ablation-study}, PSDF can reconstruct high-fidelity surfaces owing to the importance-guided rendering can force the model  focus more on the color learning of surface intersection points.

\begin{figure}[htbp]
	\setlength{\abovecaptionskip}{1.5mm}
	\setlength{\belowcaptionskip}{0cm}
	\center
	\includegraphics[width=0.48\textwidth]{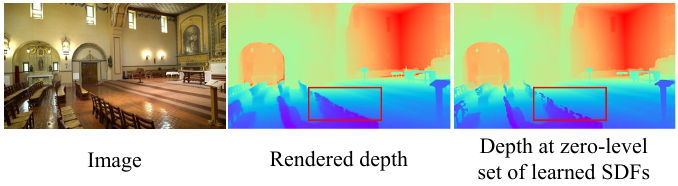}
	\caption{Visual comparison between the depth map from volume rendering and the depth map at the zero-level set of learned SDFs.}
	\label{IPIR}
\end{figure}

In addition, Fig. \ref{IPIR} presents a visual comparison between the depth extracted via volume rendering and the depth derived from the zero-level set of learned SDFs. As shown, although the depth at the zero-level set of learned SDFs corresponds to the point on the extracted surface, there are missing depth in certain regions.  Conversely, the rendered depth is more complete.  
Therefore, instead of solely using the depth at the zero-level set of learned SDFs for importance rendering, both rendered depth and depth at the zero-level set are used for importance rendering.

\begin{table}[htbp]
\caption{Ablation experimental results for other components used in Baseline model on ``Church" and ``Ignatius" of Tanks and Temple dataset. (higher is better).}
\label{tab:Ablation-study-detailed}
\centering
\small
\resizebox{\linewidth}{!}{
\begin{tabular}{l|cccc|ccc}
\hline
Model & AE & $\mathcal{L}_{\textrm{smooth}}$ &$\mathcal{L}_{\textrm{normal}}$ &$\mathcal{L}_{\textrm{bias}}$ & Precision $\uparrow$ & Recall $\uparrow$ & F$_1$ $\uparrow$ \\
\hline
Model-E &  &  & &  & 8.09& 8.36&7.89  \\
Model-F & \checkmark &  & &  &20.48 &23.75 &21.22 \\
Model-G &\checkmark & \checkmark & & &23.06 & 25.76 &23.62 \\
Model-H & \checkmark & \checkmark &\checkmark & &29.76 &30.92 &29.97 \\
Baseline  & \checkmark & \checkmark  & \checkmark &\checkmark &31.80 & 31.92 & 31.6  \\
\hline
\end{tabular}}
\end{table}

\subsubsection{Other Components}

\noindent \textbf{Appearance Embedding} 
Due to complex lighting changes in real scenes, it disrupts the implicit photometric consistency inherent in NISR. By employing appearance embedding \cite{MartinBrualla2021} for each image to the color network, it allows the model to hold the implicit photometric consistency. The comparison results of Model-E and Model-F presented in Table \ref{tab:Ablation-study-detailed} show that benefited from the appearance embedding, both precision and recall are improved.

\noindent \textbf{Smooth Loss $\mathcal{L}_{\textrm{smooth}}$} 
To improve the smoothness of the recovered surface, the smooth loss \cite{Oechsle2021} is applied.
Thanks to the smoothness constraints introduced by the smooth loss, the performance of Model-G is boosted compared with that of Model-F.

\noindent \textbf{Normal Consistency Loss $\mathcal{L}_{\textrm{normal}}$} 
In addressing the challenge of photoconsistency ambiguity in regions with low-texture, the normal consistency loss \cite{Yu2022} is imposed on the optimization process of PSDF. Owing to the geometric constraints provided by $\mathcal{L}_{\textrm{normal}}$ on the learned normal, it is beneficial for the reconstruction of the plane and low-textured areas. The comparison results of Model-G and Model-H validate the effectiveness of $\mathcal{L}_{\textrm{normal}}$.

\noindent \textbf{Geometry Bias Loss $\mathcal{L}_{\textrm{bias}}$} 
To alleviate the biased surface rendering, the geometry bias loss \cite{Chen2023} is introduced. In this case, the Baseline model used in the ablation study of the main paper is formed. Thanks to its regularization of biased surface rendering, the performance of the Baseline is further improved compared with that of Model-H.

\subsection{RGB Image Synthesis} 
\begin{figure}[htbp]
	\setlength{\abovecaptionskip}{1.5mm}
	\setlength{\belowcaptionskip}{0cm}
	\center
	\includegraphics[width=0.48\textwidth]{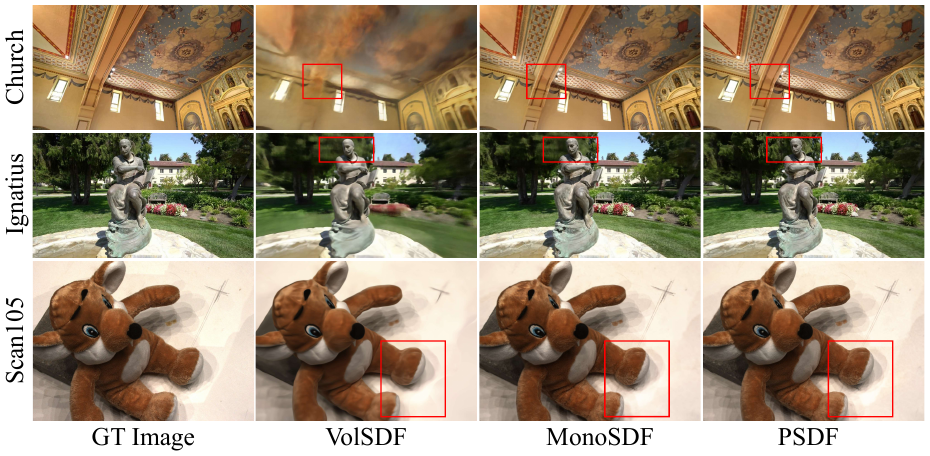}
	\caption{Visualization comparison of RGB images synthesised by various methods on  Tanks and Temples and DTU datasets.}
	\label{image_synthesis}
\end{figure}

The appearance embedding \cite{MartinBrualla2021} is used to handle the complex lighting changes between different views on the Tanks and Temples dataset, which learns a corresponding real-valued appearance embedding vector for each image. This means that with the use of appearance embedding \cite{MartinBrualla2021}, the model can only render views that have appeared in the training views, i.e., the model can not render novel views.
The time consumption for RGB image synthesis is significant, e.g, it costs about 2.32\textit{min}, 1.47\textit{min} and 1.63\textit{min} to render an image with resolution $1024 \times 576$ on a GeForce RTX 2080Ti with 1000 pixels in a minibatch.
Moreover, this paper mainly focuses on surface mesh reconstruction. For these reasons, we do not perform a quantitative analysis of RGB image synthesis. The visualization comparison of RGB images synthesised by VolSDF \cite{Yariv2021}, MonoSDF \cite{Yu2022} and PSDF on ``Church" and ``Ignatius" of Tanks and Temples dataset and ``Scan105"  of DTU dataset are presented in Fig \ref{image_synthesis}. It can be seen that PSDF can recover more details of the scenes.

\begin{table}[htbp]	
	\centering
	\small
	\caption{Comparison results of memory consumption, run-time and the reconstruction quality on the training set of Tanks and Temples dataset. }
	\label{tab:Computational Efficiency}
	\resizebox{\linewidth}{!}{
		\begin{tabular}{c|cc|cc|c}
			\hline
			\multirow{2}{*}{Model} 
			&\multicolumn{2}{c|}{Training} & \multicolumn{2}{c|}{Inference} &TnT \\
			&Mem. (GB)& Time (\textit{h})& Mem. (GB)  & Time (\textit{s}) & $F_1$ (\%)\\
			\hline
			VolSDF \cite{Yariv2021} & 7.91 & 14.88 & 2.15 & 22.33 & 11.73\\
			MonoSDF \cite{Yu2022} & 6.82 & 11.50 & 2.83 & 13.31 & 30.32\\
			Neuralangelo \cite{Li2023} & 18.61 &22.41 & 13.42 & 13.43 & 50\\
			PSDF & 8.11 & 17.05 & 2.83 & 13.66 &53.68\\
			\hline
	\end{tabular}}
\end{table}

\subsection{Memory and Run-time Comparison} 
Table \ref{tab:Computational Efficiency} presents comparison results of memory consumption, run-time and the reconstruction quality on the training set of Tanks and Temples dataset. The memory consumption and run-time are conducted on a outdoor scene of Tanks and Temples (TnT) using a GeForce RTX 3090 where the inference is for the surface with resolution $512^3$. It can be seen from Table \ref{tab:Computational Efficiency}, PSDF achieves a good balance between computational efficiency and reconstruction performance.

\subsection{Limitations}
\begin{figure}[htbp]
	\setlength{\abovecaptionskip}{1.5mm}
	\setlength{\belowcaptionskip}{0cm}
	\center
	\includegraphics[width=0.42\textwidth]{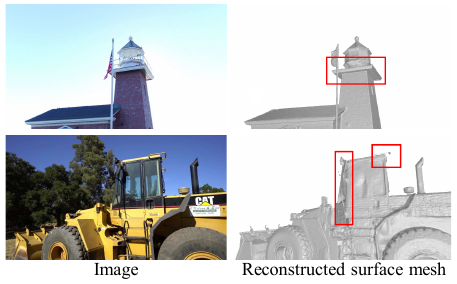}
	\caption{Failure cases. PSDF fails to recover thin structures.}
	\label{failed_case}
\end{figure}

As a NISR method, PSDF inherits its characteristic of requiring per-scene optimization. How to accelerate training is a direction worth exploring in the future. 
Moreover, as depicted in Fig. \ref{failed_case}, the PSDF fails to recover the thin structure. 
This issue arises due to the limited occurrence of pixels representing these delicate structures across the entire image. This resulting low probability of these pixels being sampled hampers the optimization process for this particular region.
Furthermore, the images from the Tanks and Temples dataset are downsampled to the resolution with $1024 \times 576$. To mitigate this challenge, several strategies can be explored during the training phase in the future, such as increasing the number of sampled pixels in a batch and improving the resolution of the image, as well as guiding pixel sampling through edge information.

\section{Conclusion}
In this paper, we introduce a novel framework for multi-view reconstruction called PSDF, which leverages both external and internal priors to achieve high-quality surface reconstruction in complex uncontrolled scenes. 
To impose a robust photoconsistency for powerful geometric consistency constrain, the visibility-aware feature consistency loss is introduce by leveraging the image features and visibility information obtained from a pretrained MVS network.
Based on the depth prior estimated by the pretrained MVS network, the depth-assisted sampling is presented to efficient locating the surface intersection points.
Furthermore, we exploit the inherent prior of NISR to alleviate biased surface rendering through the proposed internal prior-guided importance rendering.
Extensive experiments on Tanks and Temples and DTU datasets substantiate the excellent performance of PSDF. Our approach excels not only in simple scenes but also in complex scenes.


%

%

\ifCLASSOPTIONcompsoc
  \section*{Acknowledgments}
\else
  \section*{Acknowledgment}
\fi

This work was supported by the National Natural Science Foundation of China under Grants 62176096 and 61991412.

\ifCLASSOPTIONcaptionsoff
  \newpage
\fi



%

\bibliographystyle{IEEEtran}
\bibliography{IEEEabrv}{}

\begin{thebibliography}{10}
\providecommand{\url}[1]{#1}
\csname url@samestyle\endcsname
\providecommand{\newblock}{\relax}
\providecommand{\bibinfo}[2]{#2}
\providecommand{\BIBentrySTDinterwordspacing}{\spaceskip=0pt\relax}
\providecommand{\BIBentryALTinterwordstretchfactor}{4}
\providecommand{\BIBentryALTinterwordspacing}{\spaceskip=\fontdimen2\font plus
\BIBentryALTinterwordstretchfactor\fontdimen3\font minus
  \fontdimen4\font\relax}
\providecommand{\BIBforeignlanguage}[2]{{%
\expandafter\ifx\csname l@#1\endcsname\relax
\typeout{** WARNING: IEEEtran.bst: No hyphenation pattern has been}%
\typeout{** loaded for the language `#1'. Using the pattern for}%
\typeout{** the default language instead.}%
\else
\language=\csname l@#1\endcsname
\fi
#2}}
\providecommand{\BIBdecl}{\relax}
\BIBdecl

\bibitem{Li2022}
H.~Li, X.~Yang, H.~Zhai, Y.~Liu, H.~Bao, and G.~Zhang, ``Vox-surf: Voxel-based
  implicit surface representation,'' \emph{IEEE Transactions on Visualization
  and Computer Graphics}, 2022.

\bibitem{Petrov2023}
D.~Petrov, M.~Gadelha, R.~M{\v{e}}ch, and E.~Kalogerakis, ``Anise:
  Assembly-based neural implicit surface reconstruction,'' \emph{IEEE
  Transactions on Visualization and Computer Graphics}, 2023.

\bibitem{Schoenberger2016}
J.~L. Sch{\"o}nberger, E.~Zheng, J.-M. Frahm, and M.~Pollefeys, ``Pixelwise
  view selection for unstructured multi-view stereo,'' in \emph{European
  Conference on Computer Vision}.\hskip 1em plus 0.5em minus 0.4em\relax
  Springer, 2016, pp. 501--518.

\bibitem{Wei2022}
Z.~Wei, Q.~Zhu, C.~Min, Y.~Chen, and G.~Wang, ``Bidirectional hybrid lstm based
  recurrent neural network for multi-view stereo,'' \emph{IEEE Transactions on
  Visualization and Computer Graphics}, 2022.

\bibitem{Zhang2022}
J.~Zhang, Y.~Yao, S.~Li, T.~Fang, D.~McKinnon, Y.~Tsin, and L.~Quan, ``Critical
  regularizations for neural surface reconstruction in the wild,'' in
  \emph{Proceedings of the IEEE/CVF Conference on Computer Vision and Pattern
  Recognition}, 2022, pp. 6270--6279.

\bibitem{Wang2021}
P.~Wang, L.~Liu, Y.~Liu, C.~Theobalt, T.~Komura, and W.~Wang, ``Neus: Learning
  neural implicit surfaces by volume rendering for multi-view reconstruction,''
  in \emph{Advances in Neural Information Processing Systems}, vol.~34, 2021,
  pp. 27\,171--27\,183.

\bibitem{Yariv2021}
L.~Yariv, J.~Gu, Y.~Kasten, and Y.~Lipman, ``Volume rendering of neural
  implicit surfaces,'' in \emph{Advances in Neural Information Processing
  Systems}, vol.~34, 2021, pp. 4805--4815.

\bibitem{Yariv2020}
L.~Yariv, Y.~Kasten, D.~Moran, M.~Galun, M.~Atzmon, B.~Ronen, and Y.~Lipman,
  ``Multiview neural surface reconstruction by disentangling geometry and
  appearance,'' \emph{Advances in Neural Information Processing Systems},
  vol.~33, pp. 2492--2502, 2020.

\bibitem{Mildenhall2020}
B.~Mildenhall, P.~P. Srinivasan, M.~Tancik, J.~T. Barron, R.~Ramamoorthi, and
  R.~Ng, ``Nerf: Representing scenes as neural radiance fields for view
  synthesis,'' in \emph{European Conference on Computer Vision}.\hskip 1em plus
  0.5em minus 0.4em\relax Springer, 2020, pp. 405--421.

\bibitem{Yu2022}
Z.~Yu, S.~Peng, M.~Niemeyer, T.~Sattler, and A.~Geiger, ``Monosdf: Exploring
  monocular geometric cues for neural implicit surface reconstruction,'' in
  \emph{Advances in Neural Information Processing Systems}, 2022.

\bibitem{Wang2022a}
J.~Wang, P.~Wang, X.~Long, C.~Theobalt, T.~Komura, L.~Liu, and W.~Wang,
  ``Neuris: Neural reconstruction of indoor scenes using normal priors,'' in
  \emph{Computer Vision--ECCV 2022: 17th European Conference, Tel Aviv, Israel,
  October 23--27, 2022, Proceedings, Part XXXII}.\hskip 1em plus 0.5em minus
  0.4em\relax Springer, 2022, pp. 139--155.

\bibitem{Fu2022}
Q.~Fu, Q.~Xu, Y.~S. Ong, and W.~Tao, ``Geo-neus: Geometry-consistent neural
  implicit surfaces learning for multi-view reconstruction,'' in \emph{Advances
  in Neural Information Processing Systems}, vol.~35, 2022, pp. 3403--3416.

\bibitem{Zhang2021}
J.~Zhang, Y.~Yao, and L.~Quan, ``Learning signed distance field for multi-view
  surface reconstruction,'' in \emph{Proceedings of the IEEE/CVF International
  Conference on Computer Vision}, 2021, pp. 6525--6534.

\bibitem{Darmon2022}
F.~Darmon, B.~Bascle, J.-C. Devaux, P.~Monasse, and M.~Aubry, ``Improving
  neural implicit surfaces geometry with patch warping,'' in \emph{Proceedings
  of the IEEE/CVF Conference on Computer Vision and Pattern Recognition}, 2022,
  pp. 6260--6269.

\bibitem{Chen2023}
D.~Chen, P.~Zhang, I.~Feldmann, O.~Schreer, and P.~Eisert, ``Recovering fine
  details for neural implicit surface reconstruction,'' in \emph{Proceedings of
  the IEEE/CVF Winter Conference on Applications of Computer Vision}, 2023, pp.
  4330--4339.

\bibitem{Zhang2023}
Y.~Zhang, Z.~Hu, H.~Wu, M.~Zhao, L.~Li, Z.~Zou, and C.~Fan, ``Towards unbiased
  volume rendering of neural implicit surfaces with geometry priors,'' in
  \emph{Proceedings of the IEEE/CVF Conference on Computer Vision and Pattern
  Recognition}, 2023, pp. 4359--4368.

\bibitem{Yao2018}
Y.~Yao, Z.~Luo, S.~Li, T.~Fang, and L.~Quan, ``Mvsnet: Depth inference for
  unstructured multi-view stereo,'' in \emph{Proceedings of the European
  conference on computer vision (ECCV)}, 2018, pp. 767--783.

\bibitem{Knapitsch2017}
A.~Knapitsch, J.~Park, Q.-Y. Zhou, and V.~Koltun, ``Tanks and temples:
  Benchmarking large-scale scene reconstruction,'' \emph{ACM Transactions on
  Graphics (ToG)}, vol.~36, no.~4, pp. 1--13, 2017.

\bibitem{Aanaes2016}
H.~Aan{\ae}s, R.~R. Jensen, G.~Vogiatzis, E.~Tola, and A.~B. Dahl,
  ``Large-scale data for multiple-view stereopsis,'' \emph{International
  Journal of Computer Vision}, vol. 120, pp. 153--168, 2016.

\bibitem{Vu2011}
H.-H. Vu, P.~Labatut, J.-P. Pons, and R.~Keriven, ``High accuracy and
  visibility-consistent dense multiview stereo,'' \emph{IEEE transactions on
  pattern analysis and machine intelligence}, vol.~34, no.~5, pp. 889--901,
  2011.

\bibitem{Kazhdan2013}
M.~Kazhdan and H.~Hoppe, ``Screened poisson surface reconstruction,'' \emph{ACM
  Transactions on Graphics (ToG)}, vol.~32, no.~3, pp. 1--13, 2013.

\bibitem{Curless1996}
B.~Curless and M.~Levoy, ``A volumetric method for building complex models from
  range images,'' in \emph{Proceedings of the 23rd annual conference on
  Computer graphics and interactive techniques}, 1996, pp. 303--312.

\bibitem{Gu2020}
X.~Gu, Z.~Fan, S.~Zhu, Z.~Dai, F.~Tan, and P.~Tan, ``Cascade cost volume for
  high-resolution multi-view stereo and stereo matching,'' in \emph{Proceedings
  of the IEEE/CVF conference on computer vision and pattern recognition}, 2020,
  pp. 2495--2504.

\bibitem{Su2023}
W.~Su and W.~Tao, ``Efficient edge-preserving multi-view stereo network for
  depth estimation,'' in \emph{Proceedings of the AAAI Conference on Artificial
  Intelligence}, vol.~37, no.~2, 2023, pp. 2348--2356.

\bibitem{Zhang2023a}
Z.~Zhang, R.~Peng, Y.~Hu, and R.~Wang, ``Geomvsnet: Learning multi-view stereo
  with geometry perception,'' in \emph{Proceedings of the IEEE/CVF Conference
  on Computer Vision and Pattern Recognition}, 2023, pp. 21\,508--21\,518.

\bibitem{Niemeyer2020}
M.~Niemeyer, L.~Mescheder, M.~Oechsle, and A.~Geiger, ``Differentiable
  volumetric rendering: Learning implicit 3d representations without 3d
  supervision,'' in \emph{Proceedings of the IEEE/CVF Conference on Computer
  Vision and Pattern Recognition}, 2020, pp. 3504--3515.

\bibitem{Oechsle2021}
M.~Oechsle, S.~Peng, and A.~Geiger, ``Unisurf: Unifying neural implicit
  surfaces and radiance fields for multi-view reconstruction,'' in
  \emph{Proceedings of the IEEE/CVF International Conference on Computer
  Vision}, 2021, pp. 5589--5599.

\bibitem{Mueller2022}
T.~M{\"u}ller, A.~Evans, C.~Schied, and A.~Keller, ``Instant neural graphics
  primitives with a multiresolution hash encoding,'' \emph{ACM Transactions on
  Graphics (ToG)}, vol.~41, no.~4, pp. 1--15, 2022.

\bibitem{Cai2023}
B.~Cai, J.~Huang, R.~Jia, C.~Lv, and H.~Fu, ``Neuda: Neural deformable anchor
  for high-fidelity implicit surface reconstruction,'' in \emph{Proceedings of
  the IEEE/CVF Conference on Computer Vision and Pattern Recognition}, 2023,
  pp. 8476--8485.

\bibitem{Rosu2023}
R.~A. Rosu and S.~Behnke, ``Permutosdf: Fast multi-view reconstruction with
  implicit surfaces using permutohedral lattices,'' in \emph{Proceedings of the
  IEEE/CVF Conference on Computer Vision and Pattern Recognition}, 2023, pp.
  8466--8475.

\bibitem{Li2023}
Z.~Li, T.~M{\"u}ller, A.~Evans, R.~H. Taylor, M.~Unberath, M.-Y. Liu, and C.-H.
  Lin, ``Neuralangelo: High-fidelity neural surface reconstruction,'' in
  \emph{Proceedings of the IEEE/CVF Conference on Computer Vision and Pattern
  Recognition}, 2023, pp. 8456--8465.

\bibitem{Guo2022}
H.~Guo, S.~Peng, H.~Lin, Q.~Wang, G.~Zhang, H.~Bao, and X.~Zhou, ``Neural 3d
  scene reconstruction with the manhattan-world assumption,'' in
  \emph{Proceedings of the IEEE/CVF Conference on Computer Vision and Pattern
  Recognition}, 2022, pp. 5511--5520.

\bibitem{Eftekhar2021}
A.~Eftekhar, A.~Sax, J.~Malik, and A.~Zamir, ``Omnidata: A scalable pipeline
  for making multi-task mid-level vision datasets from 3d scans,'' in
  \emph{Proceedings of the IEEE/CVF International Conference on Computer
  Vision}, 2021, pp. 10\,786--10\,796.

\bibitem{Hartley2003}
R.~Hartley and A.~Zisserman, \emph{Multiple view geometry in computer
  vision}.\hskip 1em plus 0.5em minus 0.4em\relax Cambridge university press,
  2003.

\bibitem{Paszke2019}
A.~Paszke, S.~Gross, F.~Massa, A.~Lerer, J.~Bradbury, G.~Chanan, T.~Killeen,
  Z.~Lin, N.~Gimelshein, L.~Antiga \emph{et~al.}, ``Pytorch: An imperative
  style, high-performance deep learning library,'' \emph{Advances in neural
  information processing systems}, vol.~32, 2019.

\bibitem{Kingma2014}
D.~P. Kingma and J.~Ba, ``Adam: A method for stochastic optimization,''
  \emph{arXiv preprint arXiv:1412.6980}, 2014.

\bibitem{Barron2022}
J.~T. Barron, B.~Mildenhall, D.~Verbin, P.~P. Srinivasan, and P.~Hedman,
  ``Mip-nerf 360: Unbounded anti-aliased neural radiance fields,'' in
  \emph{Proceedings of the IEEE/CVF Conference on Computer Vision and Pattern
  Recognition}, 2022, pp. 5470--5479.

\bibitem{Atzmon2020}
M.~Atzmon and Y.~Lipman, ``Sal: Sign agnostic learning of shapes from raw
  data,'' in \emph{Proceedings of the IEEE/CVF Conference on Computer Vision
  and Pattern Recognition}, 2020, pp. 2565--2574.

\bibitem{MartinBrualla2021}
R.~Martin-Brualla, N.~Radwan, M.~S. Sajjadi, J.~T. Barron, A.~Dosovitskiy, and
  D.~Duckworth, ``Nerf in the wild: Neural radiance fields for unconstrained
  photo collections,'' in \emph{Proceedings of the IEEE/CVF Conference on
  Computer Vision and Pattern Recognition}, 2021, pp. 7210--7219.

\bibitem{William1987}
E.~L. William and E.~C. Harvey, ``Marching cubes: A high resolution 3d surface
  construction algorithm,'' \emph{ACM SIGGRAPH Computer Graphics}, vol.~21,
  no.~4, pp. 163--169, 1987.

\bibitem{Yao2020}
Y.~Yao, Z.~Luo, S.~Li, J.~Zhang, Y.~Ren, L.~Zhou, T.~Fang, and L.~Quan,
  ``Blendedmvs: A large-scale dataset for generalized multi-view stereo
  networks,'' in \emph{Proceedings of the IEEE/CVF Conference on Computer
  Vision and Pattern Recognition}, 2020, pp. 1790--1799.

\bibitem{Xu2022b}
Q.~Xu, W.~Su, Y.~Qi, W.~Tao, and M.~Pollefeys, ``Learning inverse depth
  regression for pixelwise visibility-aware multi-view stereo networks,''
  \emph{International Journal of Computer Vision}, vol. 130, no.~8, pp.
  2040--2059, 2022.

\bibitem{Su2022}
W.~Su, Q.~Xu, and W.~Tao, ``Uncertainty guided multi-view stereo network for
  depth estimation,'' \emph{IEEE Transactions on Circuits and Systems for Video
  Technology}, vol.~32, no.~11, pp. 7796--7808, 2022.

\bibitem{Wang2021a}
F.~Wang, S.~Galliani, C.~Vogel, P.~Speciale, and M.~Pollefeys, ``Patchmatchnet:
  Learned multi-view patchmatch stereo,'' in \emph{Proceedings of the IEEE/CVF
  conference on computer vision and pattern recognition}, 2021, pp.
  14\,194--14\,203.

\bibitem{Zhang2020}
K.~Zhang, G.~Riegler, N.~Snavely, and V.~Koltun, ``Nerf++: Analyzing and
  improving neural radiance fields,'' \emph{arXiv preprint arXiv:2010.07492},
  2020.

\bibitem{Lin2017}
T.-Y. Lin, P.~Doll{\'a}r, R.~Girshick, K.~He, B.~Hariharan, and S.~Belongie,
  ``Feature pyramid networks for object detection,'' in \emph{Proceedings of
  the IEEE conference on computer vision and pattern recognition}, 2017, pp.
  2117--2125.

\end{thebibliography}

%








\end{document}